\documentclass[twoside]{article}

\usepackage[utf8]{inputenc} 
\usepackage[T1]{fontenc}    
\usepackage{hyperref}       
\usepackage{url}            
\usepackage{booktabs}       
\usepackage{amsfonts}       
\usepackage{nicefrac}       
\usepackage{microtype}      
\usepackage{graphicx}
\usepackage{amsmath}
\usepackage{subcaption}
\usepackage[ruled,vlined]{algorithm2e}

\newcommand{\mtx}{}

\def\E{\mathbb{E}}

\def\name{VAELLS}
\def\namelong{Variational Autoencoder with Learned Latent Structure}

\newcommand{\abs}[1]{\left\lvert#1\right\rvert}
\newcommand{\norm}[1]{\left\lVert#1\right\rVert}
\newcommand{\argmin}{\mathop{\mathrm{argmin}}}    
\newcommand{\argmax}{\mathop{\mathrm{argmax}}}

\usepackage{hyperref}

\usepackage[accepted]{aistats2021}
\usepackage[round]{natbib}

\begin{document}

%

%

\twocolumn[

\aistatstitle{Variational Autoencoder with Learned Latent Structure}

\aistatsauthor{ Marissa Connor \And Gregory Canal \And Christopher Rozell }

\aistatsaddress{ School of Electrical and Computer Engineering\\
  Georgia Institute of Technology\\
  Atlanta, GA 30332 \\
  \texttt{(marissa.connor,gregory.canal,crozell)@gatech.edu} } ]

\begin{abstract}
  The manifold hypothesis states that high-dimensional data can be modeled as lying on or near a low-dimensional, nonlinear manifold. Variational Autoencoders (VAEs) approximate this manifold by learning mappings from low-dimensional latent vectors to high-dimensional data while encouraging a global structure in the latent space through the use of a specified prior distribution. When this prior does not match the structure of the true data manifold, it can lead to a less accurate model of the data. To resolve this mismatch, we introduce the \namelong~(\name) which incorporates a learnable manifold model into the latent space of a VAE. This enables us to learn the nonlinear manifold structure from the data and use that structure to define a prior in the latent space. The integration of a latent manifold model not only ensures that our prior is well-matched to the data, but also allows us to define generative transformation paths in the latent space and describe class manifolds with transformations stemming from examples of each class. We validate our model on examples with known latent structure and also demonstrate its capabilities on a real-world dataset.\footnote{Code is available at \url{https://github.com/siplab-gt/VAELLS}.}
\end{abstract}

\section{INTRODUCTION}

Generative models represent complex data distributions by defining generator functions that map low-dimensional latent vectors to high-dimensional data outputs. In particular, generative models such as variational autoencoders (VAEs)~\citep{kingma2013auto,rezende2014stochastic} and generative adversarial networks~\citep{goodfellow2014generative} sample from a latent space with a specified prior distribution in order to generate new, realistic samples. VAEs have the additional benefit of an encoder that maps data inputs into a latent space. This latent space embedding of data points makes VAEs an effective tool for generating and understanding variations in natural data.

According to the manifold hypothesis, high-dimensional data can often be modeled as lying on or near a low-dimensional, nonlinear manifold~\citep{fefferman2016testing}. There are many manifold learning techniques that compute embeddings of high-dimensional data but very few of them have the ability to generate new points on the manifold~\citep{tenenbaum2000global,roweis2000nonlinear,dollar2007learning,bengio2005non}. Meanwhile, VAEs can both embed high-dimensional data in a low-dimensional space and generate outputs from that space,  making them a convenient model for representing low-dimensional data manifolds. 

However, there are several aspects of the traditional VAE framework that prevent it from faithfully representing complex natural variations on manifolds associated with separate data classes. First, VAEs enforce a global structure in the latent space through the use of a prior distribution, and that prior may not match the true data manifold; this model mismatch can result in a less accurate generative model of the data. Second, natural paths, which interpolate or extrapolate natural data variations, are poorly defined in the latent space of traditional VAEs. In many cases, the data transformations are defined using linear paths in a Euclidean latent space~\citep{radford2015unsupervised}. These simple paths can diverge from the true data manifold, leading to interpolated points that result in unrealistic decoded image outputs. Finally, traditional VAEs encourage points from all data classes to cluster around the origin in the latent space~\citep{kingma2013auto}. Without adequate class separation, traversing the latent space can easily result in a change of class, making it difficult to learn identity-preserving transformations and subsequently use them to understand within-class relationships.

In this paper we incorporate a generative manifold model known as \emph{transport operators} \citep{culpepper2009learning} into the latent space of a VAE, enabling us to learn the manifold structure from the data and use that structure to define an appropriate prior in the latent space. This approach not only ensures that the prior is well-matched to the data, but it also allows us to define nonlinear transformation paths in the latent space and describe class manifolds with transformations stemming from examples of each class. Our model, named \namelong~(\name), more effectively represents natural data manifolds than existing techniques, leads to more accurate generative data outputs, and results in a richer understanding of data variations.

\section{METHODS}

\subsection{Transport Operators}
\label{sec:TO}

The exact structure of natural data manifolds is typically unknown and manifold learning techniques have been introduced to discover the low dimensional structure of data. While there are a variety of manifold learning techniques~\citep{tenenbaum2000global,roweis2000nonlinear,dollar2007learning,bengio2005non}, we desire a manifold model that allows us to learn the data structure, generate points outside of the training set, and map out smooth manifold paths through the latent space.  The transport operator manifold model~\citep{culpepper2009learning} satisfies these requirements by defining a manifold using learned operators that traverse the low-dimensional manifold surface.

Specifically, the basis of the transport operator approach is the linear dynamical system model $\dot{z}=\mtx{A}z$, which defines the dynamics of point $z \in \mathbb{R}^d$ through $\mtx{A} \in \mathbb{R}^{d \times d}$. The solution to this differential equation defines a temporal path given by $z_t =  \mathrm{expm}(\mtx{A}t)z_0$, where $\mathrm{expm}$ is the matrix exponential. This path definition can be generalized to define the transformation between any two points $z_0,z_1 \in \mathbb{R}^d$ on a low-dimensional natural data manifold without an explicit time component by defining
$
z_1 = \mathrm{expm}(\mtx{A})z_0 + n
$,
where $n$ is white Gaussian noise. To allow for different geometrical characteristics at various points on the manifold, our model should have the flexibility to define a different dynamics matrix $\mtx{A}$ between each pair of points. The transport operator technique achieves this property by defining a dynamics matrix that can be decomposed as a weighted sum of $M$ transport operator dictionary elements ($\mtx{\Psi}_m \in \mathbb{R}^{d\times d}$):
\begin{equation} 
\mtx{A} = \sum_{m=1}^M{\mtx{\Psi}_m c_m}.
\end{equation}
The transport operators $\{\mtx{\Psi}_m\}$ constitute a set of primitives that describe local characteristics over the entire manifold, while for each pair of points (i.e., at each manifold location) the geometry is governed by a small subset of operators through coefficients $c \in \mathbb{R}^M$ specific to each pair. 

The final generative manifold model is developed by incorporating the decomposable transformation matrix into the linear dynamical system model and imposing a sparsity-inducing Laplace prior on the coefficients. This Laplace prior encourages the manifold geometry between any given pair of points to be represented by only a small subset of dictionary elements. 
\begin{equation}
\begin{split}
z_1 = \mathrm{expm}\left(\sum_{m=1}^M{\mtx{\Psi}_m c_m}\right)z_0 + n \label{eq:transOptGen} \\
n \sim \mathcal{N}(0,I) \quad  c_m \sim \text{Laplace}\left(0,b\right).
\end{split}
\end{equation}

Following the unsupervised algorithm in~\citet{culpepper2009learning}, the transport operators can be learned from pairs of points on the same manifold using descent techniques that alternate between inferring the coefficients and updating the transport operators. The manifold structure can be represented by a combination of these learned transport operators and samples that are known to exist on the manifold. The transport operators constrain motion to only paths along the learned manifold and labeled samples provide starting points from which the operators can generate new samples on the manifold. In this paper we incorporate this structured model into a VAE to encourage the latent space structure to adapt to the true data manifold.

\subsection{Variational Autoencoder}

The VAE model learns a low-dimensional latent representation by defining a generator function $g : \mathcal{Z} \rightarrow \mathcal{X}$ that maps latent points $z \in \mathbb{R}^d$ to high-dimensional data points $x \in \mathbb{R}^D$. The desired objective for training a VAE is maximizing the log-likelihood of a dataset $X = \{x_1,\ldots,x_N\}$ given by $\frac{1}{N}\log p(X) = \frac{1}{N}\sum_{i=1}^{N} \log \int p(x_i,z)dz$. However, this objective is difficult to maximize, especially when parameterized by a neural network. To address this complication, VAEs instead maximize the Evidence Lower Bound (ELBO) of the marginal likelihood of each datapoint $x_i$:
\begin{equation}
    \begin{split}
    \log p(x_i) \geq \mathcal{L}(x_i) = \E_{z \sim q_{\phi}(z\mid x_i)}[-\log q_\phi(z\mid x_i) \\ + \log p_\theta(x_i,z)],
\end{split}
\end{equation}
where $q_\phi(z\mid x)$ is a variational approximation of the true posterior, parameterized by $\phi$. In the VAE neural network model, $\phi$ represents the weights of an encoder network $f_\phi(x)$.

\citet{kingma2013auto} developed an efficient method to approximate the ELBO by introducing the \textit{reparameterization trick} that enables the stochastic latent variable $z$ to be represented by a deterministic function $z = h_\phi(x,\varepsilon)$, where $\varepsilon$ is an auxiliary random variable with a parameter-free distribution. In the traditional VAE framework, the variational posterior is selected to be a multivariate Gaussian distribution, meaning that $z$ is reparameterized around the encoded point $z = f_\phi(x) + \sigma\varepsilon$ where $\varepsilon \sim \mathcal{N}(0,I)$. Additionally, the prior $p_\theta(z)$ is modeled as a zero-mean isotropic normal distribution that encourages the clustering of latent points around the origin. 

\subsection{\namelong}
\label{sec:TOVAE}
In \name, we fuse the versatile manifold learning capabilities of transport operators with the powerful generative modeling of VAEs. Specifically, we integrate transport operators into both the VAE variational posterior distribution and the prior in order to learn a latent probabilistic model that is adapted directly from the data manifold.

We start with the expanded ELBO from \citet{kingma2013auto}:
\begin{equation}
\begin{split}
    \mathcal{L}(x)=\E_{z \sim q_{\phi}(z\mid x)}\left[\log p_\theta(x\mid z) + \log p_\theta(z)\right. \\ - \left.\log q_\phi(z\mid x)\right] . \label{eq:ELBO}
    \end{split}
\end{equation}
For the likelihood $p_\theta(x\mid z)$, we follow prior work and choose an isotropic normal distribution with mean defined by the decoder network $g_\theta(z)$ and fixed variance $\sigma^2$, which has worked well in practice.

Our first key contribution lies in the selection of the variational posterior, which we choose as the family of manifold distributions parameterized by learned transport operators described in Section \ref{sec:TO}. Intuitively, this posterior measures the probability of vector $z$ lying on the manifold in the local neighborhood of the latent encoding of $x$ where the structure of the manifold is defined by learned transport operators. We encode the latent coordinates of $x$ with a neural network $f_\phi(\cdot)$ and then draw a sample from $q_\phi(z\mid x)$.

\begin{figure*}[t]

\centering
\begin{subfigure}[b]{0.45\textwidth}
  \centering
	{\includegraphics[width=0.98\textwidth]{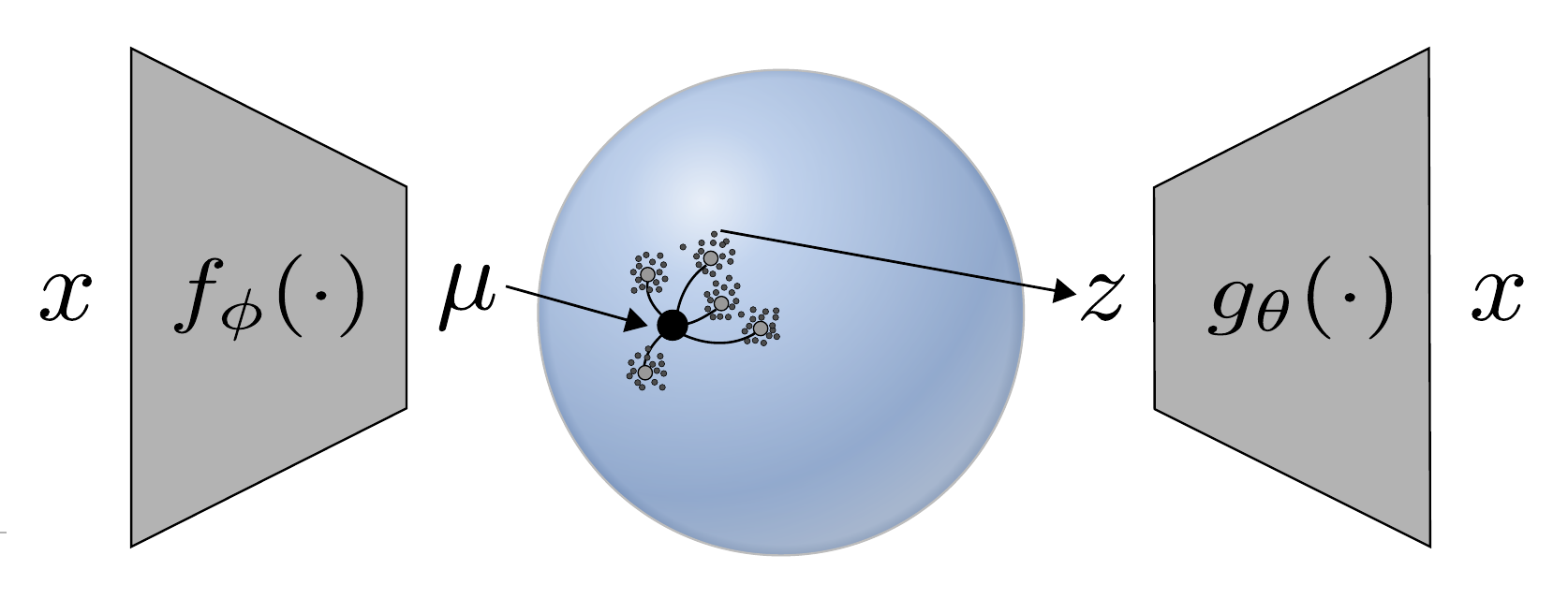}}
  \caption{}
	\label{subfig:VAELLS_encode}
\end{subfigure}
\begin{subfigure}[b]{0.10\textwidth}
  \centering
	{\includegraphics[width=0.98\textwidth]{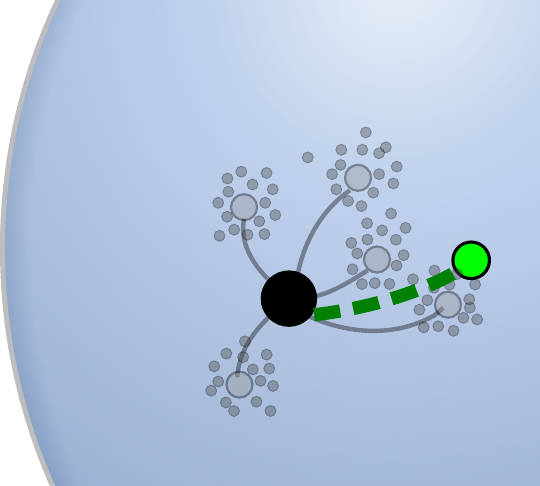}}%
	\vspace{5mm}
  \caption{}
	\label{subfig:VAELLS_post}
\end{subfigure}
\begin{subfigure}[b]{0.3\textwidth}
  \centering
	\includegraphics[width=0.98\textwidth]{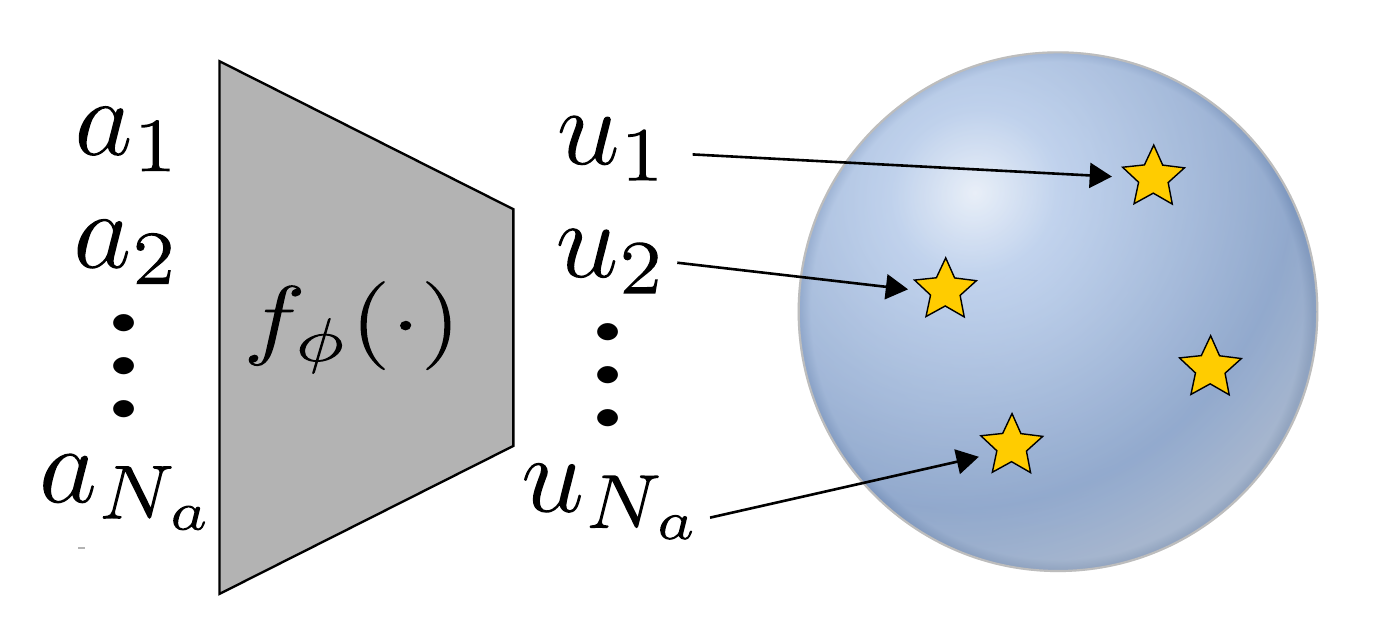}%
	\vspace{3mm}
	\caption{}
	\label{subfig:VAELLS_anchor}
\end{subfigure}
\begin{subfigure}[b]{0.12\textwidth}
  \centering
	{\includegraphics[width=0.98\textwidth]{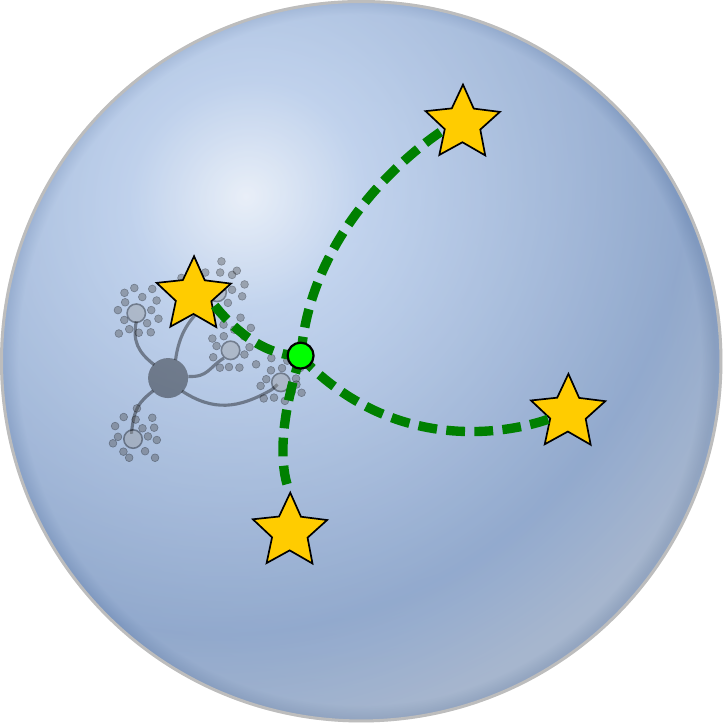}}%
	\vspace{5mm}
  \caption{}
	\label{subfig:VAELLS_prior}
\end{subfigure}

  \caption{\label{fig:VAELLS} Visualizations of the \name{}~model: (a) Posterior sampling process using transport operator-generated paths and Gaussian noise. (b) Transformation path (green dotted line) inferred between $f_\phi(x)$ (black dot) and $z$ (green dot) when computing the posterior. (c) Encoding of anchor points into the latent space. (d) Transformation paths (green dotted lines) inferred between $f_\phi(a_i)$ (yellow stars) and $z$ (green dot) when computing the prior.}
	
\end{figure*}

To approximate \eqref{eq:ELBO} with sampling, first let $L_x(z) \equiv \log p_\theta(x\mid z) + \log p_\theta(z) - \log q_\phi(z\mid x)$ and note that by marginalizing over transport operator coefficients $c$ we have $\E_{z \sim q_{\phi}(z\mid x)}\left[L_x(z)\right] = \E_{z,c\sim q_{\phi}(z,c\mid x)}\left[L_x(z)\right]$ which allows us to estimate \eqref{eq:ELBO} by sampling from $q_{\phi}(z,c\mid x)$. We draw a sample from $q_{\phi}(z,c\mid x)$ in two steps: first, as defined in the generative model in (\ref{eq:transOptGen}), we sample a set of coefficients $\widehat{c}$  from a factorized Laplace distribution $q(c)$, and then sample $z$ from $q_\phi(z \mid \widehat{c},x)$. Both of these sampling steps can be achieved with deterministic mappings on parameter-free random variates, allowing for the use of the reparameterization trick. Specifically,
\begin{equation}
\begin{split}
    \widehat{c} = l(u;b)\quad u \sim ~\mathrm{Unif}\left(-\frac12,\frac12\right)^M
    \quad
    \\ z = \mtx{T}_\Psi(\widehat{c} )f_\phi(x) + \gamma\varepsilon \quad \varepsilon \sim \mathcal{N}(0,I) , \label{eq:sampling}
    \end{split}
\end{equation}
where $l(u;b)$ is a mapping described in Appendix~\ref{app:loss_func} with Laplace scale parameter $b$ and $\mtx{T}_\Psi(c) = \mathrm{expm}\left(\sum_{m=1}^M{\mtx{\Psi}_m c_m}\right)$. 

Fig.~\ref{subfig:VAELLS_encode} shows this sampling process where the data $x$ is encoded to a mean location $\mu = f_\phi(x)$. The latent vector $z$ is the result of transforming $\mu$ by $\mtx{T}_\Psi(\widehat{c})$, which moves the vector on random paths along the manifold, and adding Gaussian noise specified by $\gamma$.

The resulting transport operator variational posterior follows as
\begin{gather}
    q(c) =\left(\frac{1}{2b}\right)^{M}\prod_{m=1}^M\exp(-\frac{\abs{c_m}}{b})\notag
\\ q_\phi(z \mid c, x)\sim \mathcal{N}\left(\mtx{T}_\Psi(c)f_\phi(x),\gamma^2 I\right)\notag
\\ q_\phi(z \mid x) = \int_c q_\phi(z \mid c,x) q(c)dc \label{eq:varpost}
\\ \approx \max_c q_\phi(z \mid c, x) q(c) , \label{eq:maxapprox}
\end{gather}
where the approximation in \eqref{eq:maxapprox} is motivated by the fact that the sparsity-inducing Laplace prior on $c$ typically results in joint distributions with $z$ that are tightly peaked in coefficient space, as described in \citet{olshausen1997sparse} (see Appendix~\ref{app:loss_func} for details). This approximation is widely used in sparse dictionary learning for computational efficiency. The inferred coefficients $c^*$ that maximize $q_\phi(z \mid c, x) q(c)$ define the estimated transformation between the encoded latent coordinates and the sampled point. The dotted green line in Fig.~\ref{subfig:VAELLS_post} shows a visualization of the inferred transformation path between $f_\phi(x)$ (the black dot) and $z$ (the green dot).

Our next key contribution lies in the construction of a prior distribution learned directly from the underlying data manifold using transport operators. To gain intuition about this prior, imagine a set of $N_a$ \emph{anchor} points in the data space that correspond to samples on the desired manifold. The data manifold structure is represented by a combination of these anchor points, encoded in the latent space, and the learned transport operators that can extrapolate the manifold structure in the latent neighborhood of each of them. Fig.~\ref{subfig:VAELLS_anchor} shows a set of anchor points encoded into the latent space which represent the scaffold off of which the manifold structure is built. The prior is defined by the same probabilistic manifold model used in the variational posterior, but starting at each anchor point $a_i$ rather than $x$. This prior requires that paths be inferred between $z$ and each encoded anchor point $u_i=f_\phi(a_i)$, visualized in Fig.~\ref{subfig:VAELLS_prior} as green dotted lines. The overall prior density for $z$ is then defined as
\begin{equation}
p_\theta(z)=\frac{1}{N_a}\sum_{i=1}^{N_a} q_\phi(z\mid a_i) . \label{eq:manprior}
\end{equation}

The introduction of anchor points into the definition of the prior provides a unique opportunity for the user to define the classes within which they want to learn natural, identity-preserving transformations. If desired, the user can choose a set of anchor points, independent of class labels, to define the full data manifold and learn transformations throughout the entire dataset. If the user wants to learn identity-preserving transformations on specific class manifolds, they can separate the anchor points into classes and define the prior with respect to class-specific anchor points. Even when the prior is defined with respect to anchor points in each class individually, the set of learned transport operators is shared over all classes. Finally, if the user desires to learn a manifold representing specific transformations of individual samples, they can select anchor points that are sampled from transformation paths of specific example data points. In practice, the anchor points are initialized by either uniformly sampling the manifold space or through manual selection by a practitioner (e.g., by selecting several anchors per data class) and they can be updated throughout training. 

In \citet{tomczak2018vae}, a prior is adopted that is similarly composed of a sum of variational posterior terms, and is motivated as an approximation to an optimal prior that maximizes the standard ELBO. However, our motivation for the prior in \eqref{eq:manprior} as a direct sampling of the data manifold is novel, and our prior more directly aligns with the data manifold because it is constructed from operators that traverse the manifold itself.

The final addition to the VAELLS objective is a Frobenius-norm regularizer on the dictionary magnitudes as used in the original transport operator objective~\citep{culpepper2009learning}. This prevents the magnitudes of the dictionaries from increasing without bound and helps identify how many transport operators are necessary to represent the manifold by reducing the magnitude of operators that are not used to generate paths between samples. We define the loss function to be minimized as the approximate negative ELBO in (\ref{eq:ELBO}), restated here for convenience: 
\begin{align}
L_x(z) \equiv \log p_\theta(x\mid z) + \log p_\theta(z) - \log q_\phi(z\mid x)\notag
    \\ \mathcal{L}_{\name}(x) =-\E_{u,\varepsilon}\left[L_x(\mtx{T}_\Psi(l(u;b))f_\phi(x)+\gamma \varepsilon)\right] \notag
    \\ +  \frac{\eta}{2}\sum_{m=1}^M \lVert\Psi_m \rVert^2_F\label{eq:concise_ELBO},
\end{align}
We optimize this loss simultaneously over encoder-decoder networks $f_\phi$ and $g_\theta$, anchor points $\{a_i\}_{1:N_a}$, and transport operators $\Psi$. The implementation details of our ELBO and its optimization are described in more detail in Appendix~\ref{app:loss_func}. 

\section{RELATED WORK}

There are currently many adaptations of the original VAE that handle a subset of the limitations addressed by \name , such as learning the prior from the data, defining continuous paths in the latent space, and separating the individual class manifolds. Table~\ref{tab:VAECompare} provides a comparison of techniques which we describe in detail below.

Traditionally, the latent space prior is defined as a Gaussian distribution for model simplicity~\citep{kingma2013auto}. However this prior encourages all points to cluster around the origin which may not occur in the natural data manifold. A mismatch between the latent prior distribution of a VAE and the data manifold structure can lead to over regularization and poor data representation. Other models incorporate more complex latent structures such as hyperspheres~\citep{davidson2018hyperspherical}, tori~\citep{rey2019diffusion,falorsi2019reparameterizing}, and hyperboloids~\citep{mathieu2019continuous,nagano2019wrapped,skopek2020mixed}. These models have demonstrated their suitability for certain datasets by choosing a prior that is the best match out of a predefined set of candidates.

However, these methods are only capable of modeling a limited number of structured priors and are not able to adapt the prior to match the data manifold itself. This is a serious drawback since, in most practical cases, the latent structure of data is unlikely to easily fit a predefined prior. The variational diffusion autoencoder (VDAE) and the $\mathcal{R}$-VAE define the latent space prior directly from the data using Brownian motion on a Riemannian manifold~\citep{li2020variational,kalatzis2020variational}. The VampPrior is defined using a sum of variational posterior distributions with learnable hyperparameters~\citep{tomczak2018vae}. As noted in discussion around (\ref{eq:manprior}), this definition of a prior incorporating variational posterior terms is the same as what is used in \name{}. However, the form of the variational posteriors differ between VampPrior and \name{}\, with \name{}\ using the structured manifold model in (\ref{eq:varpost}). This affords the \name{}\ model the additional benefits noted below.

In addition to differences in how latent space structure is represented, there are various approaches for defining natural paths in this space. In the simplest case, paths in VAEs with a Euclidean latent spaces are modeled as linear paths. Some methods define geodesic transformation paths in the latent space. This can be done by incorporating a structured latent prior, like a hypersphere, on which geodesic paths are natural to compute~\citep{davidson2018hyperspherical}. Interpolated geodesic paths can also be estimated in a Euclidean latent space using an estimated Riemannian metric~\citep{kalatzis2020variational,arvanitidis2017latent,chen2017metrics,shao2018riemannian}. However, these methods are limited to defining extrapolated paths by random walks in the latent space rather than structured paths. Finally, there are other methods that lack straightforward definitions for how to compute continuous paths from one point to another~\citep{tomczak2018vae,rey2019diffusion}. 

Another limitation of most VAE models is that they do not encourage class separation and therefore generated paths often do not represent natural identity-preserving transformations on separate class data manifolds. This makes it difficult to understand the within-class relationships in the data. Some techniques encourage class separation through the choice of a prior that does not encourage data clustering~\citep{davidson2018hyperspherical} but they do not explicitly define separate class manifolds. By defining the prior structure with respect to anchor points on specified data manifolds, \name{}\ has the flexibility to define which manifolds it wants to learn transformations on. This results in a latent space structure where transformations correspond to identity-preserving variations in the data.

Two models that have notable similarities to ours are the Lie VAE~\citep{falorsi2019reparameterizing} and the Manifold Autoencoder~\citep{connor2020representing}. Both models also use Lie group representations of transformations in the latent space. The Lie VAE model encodes the data into latent variables that are elements in a Lie group which represent transformations of a reference object. This model requires the type of Lie group transformations that the network will represent (e.g., \ $SO(3)$) to be specified prior to training which may result in a model mismatch. The Manifold Autoencoder~\citep{connor2020representing} also represents data variation in a latent space using the transport operator model, but it only defines these variations in the context of a deterministic autoencoder mapping. This approach shares the motivation of representing the structured data manifold in the latent space but it lacks the fully probabilistic generative framework modeled in \name{}. Notably the development of the probabilistic framework resulted in the introduction of anchor points which are fundamental to defining the structure of the latent manifold model. The introduction of anchor points eliminates the need to jointly  select point pairs during training (as described in Section \ref{sec:TO}) and allows \name{}\ to be trained from batches of individual training points.

\begin{table*}

\centering
\caption{Comparison of VAE Techniques}
\label{tab:VAECompare}
\begin{tabular}{|| l | l | l | l||} 
 \hline
Model & Adaptive Prior & Defines Paths  & Class Separation  \\ 
 \hline
 \hline
 VAE~\citep{kingma2013auto} & No & Linear & No\\ 
 \hline
 Hyperspherical VAE~\citep{davidson2018hyperspherical} & No & Geodesic & No  \\

 \hline
$\Delta$VAE~\citep{rey2019diffusion} & No & No & No\\
 \hline
 
 VDAE~\citep{li2020variational}  & \textbf{Yes} & Linear & No\\ 
\hline
VAE with VampPrior~\citep{tomczak2018vae} & \textbf{Yes} & No & No\\
\hline
$\mathcal{R}$-VAE~\citep{kalatzis2020variational} & \textbf{Yes} & \textbf{Nonlinear} & No\\
 \hline
 Lie VAE~\citep{falorsi2019reparameterizing} & No & \textbf{Nonlinear} & No \\
 \hline
 \name ~(our approach) & \textbf{Yes} & \textbf{Nonlinear} & \textbf{Yes}\\

 \hline
\end{tabular}

\end{table*}

\section{EXPERIMENTS}\label{sec:experiments}

Our experiments highlight the strengths of \name{}: the ability to adapt the prior to the true data manifold structure, the ability to define nonlinear paths in the latent space, and the ability to separate classes by learning identity-preserving transformations within classes specified by anchor points. First, we begin with two simple datasets with known ground truth latent structures in order to validate the ability of our model to learn the true latent structure. Next, we apply \name{}\ to rotated and naturally varying MNIST digits to show that our prior can adapt to represent rotations of individual digits through a learned operator and extend to real-world data with natural transformations.


The unique characteristics of the \name{}\ variational posterior and prior lead to specific training considerations. First, as shown in (\ref{eq:maxapprox}), to compute both the variational posterior and prior distributions we must infer transformation coefficients that maximize $q_\phi(z \mid c, x) q(c)$ and $q_\phi(z \mid c, a_i) q(c)$ respectively. This involves coefficient inference between each sampled point $z$ and its neural network encoding $f_{\phi}(x)$ as well as between $z$ and all encoded anchor points $f_{\phi}(a_i)$. The coefficient inference is performed using a conjugate gradient descent optimization solver.

For training the networks weights we use the Adam algorithm~\citep{kingma2014adam}. To add stability and improve efficiency of training, we alternate between steps where we update the network and anchor points while keeping the transport operators fixed, and steps where we update the transport operators while keeping the network weights and anchor points fixed. 

The selection of anchor points is important for learning identity-preserving transport operators. Training \name{}\ on a dataset with multiple classes requires class labels for both the anchor points and training points in order to compare each training sample with only anchor points from the same class. Anchor points are initialized by selecting training samples from each class in the input space. While we do allow for updates to the anchor points, in practice they only vary by a negligible amount during the entire training process.  Details of the network architectures and training parameters for each experiment are available in the Appendix as well as an algorithmic view of the training procedure (Algorithm 1). 

\begin{figure*}[h]

\centering
\begin{subfigure}[b]{0.17\textwidth}
  \centering
	{\includegraphics[width=0.75\textwidth]{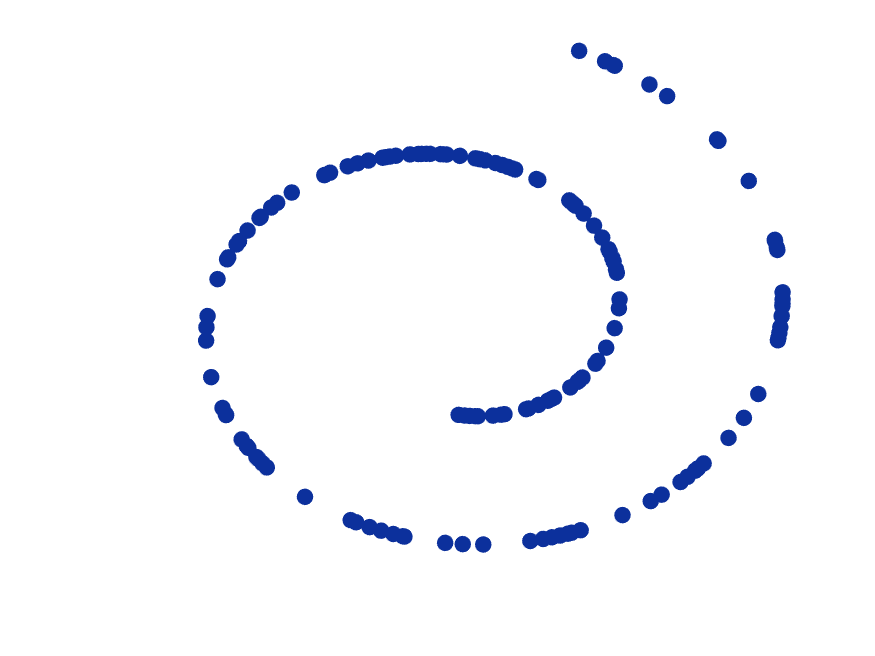}}
  \caption{}
	\label{subfig:swissGT}
\end{subfigure}
\begin{subfigure}[b]{0.19\textwidth}
  \centering
	{\includegraphics[width=0.75\textwidth]{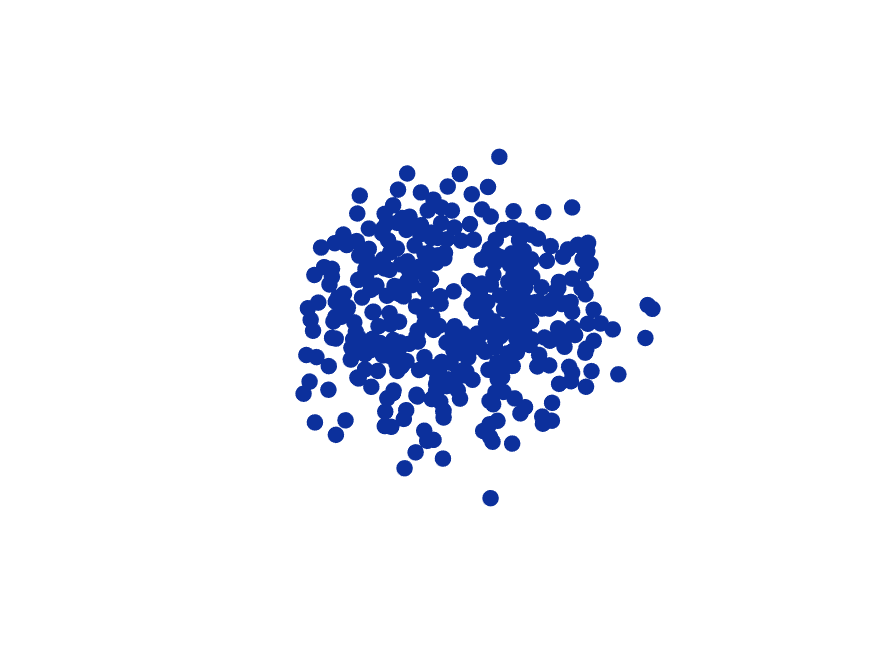}}
  \caption{}
	\label{subfig:swissVAE}
\end{subfigure}
\begin{subfigure}[b]{0.19\textwidth}
  \centering
	\includegraphics[width=0.75\textwidth]{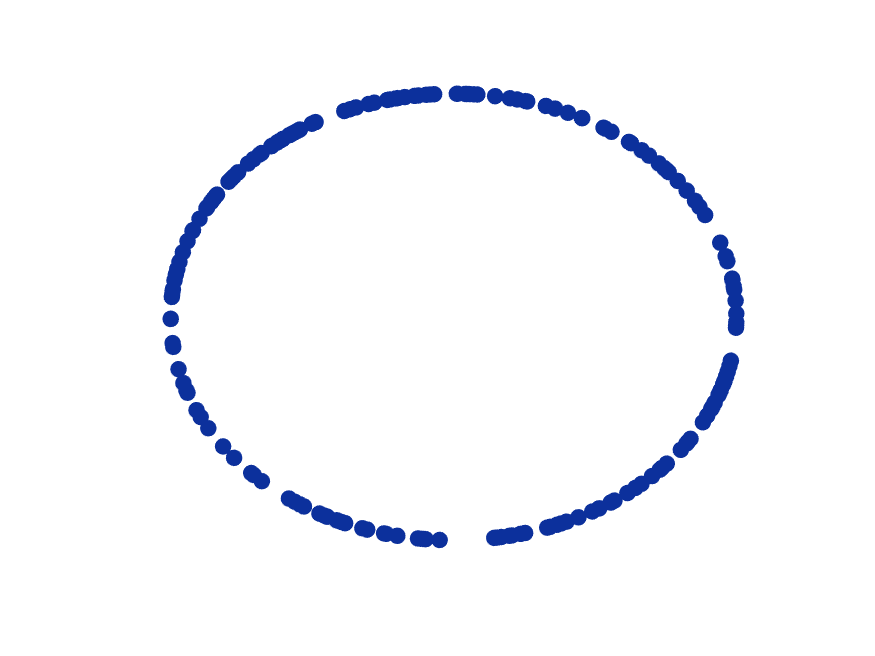}
	\caption{}
	\label{subfig:swissHVAE}
\end{subfigure}
\begin{subfigure}[b]{0.19\textwidth}
  \centering
	{\includegraphics[width=0.75\textwidth]{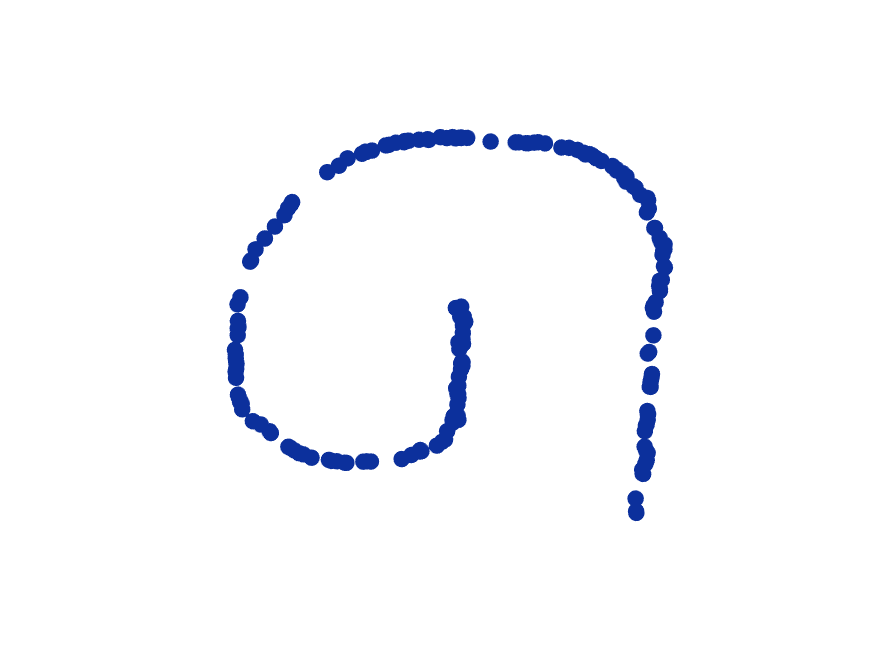}}
  \caption{}
	\label{subfig:swissVampprior}
\end{subfigure}
\begin{subfigure}[b]{0.19\textwidth}
  \centering
	{\includegraphics[width=0.75\textwidth]{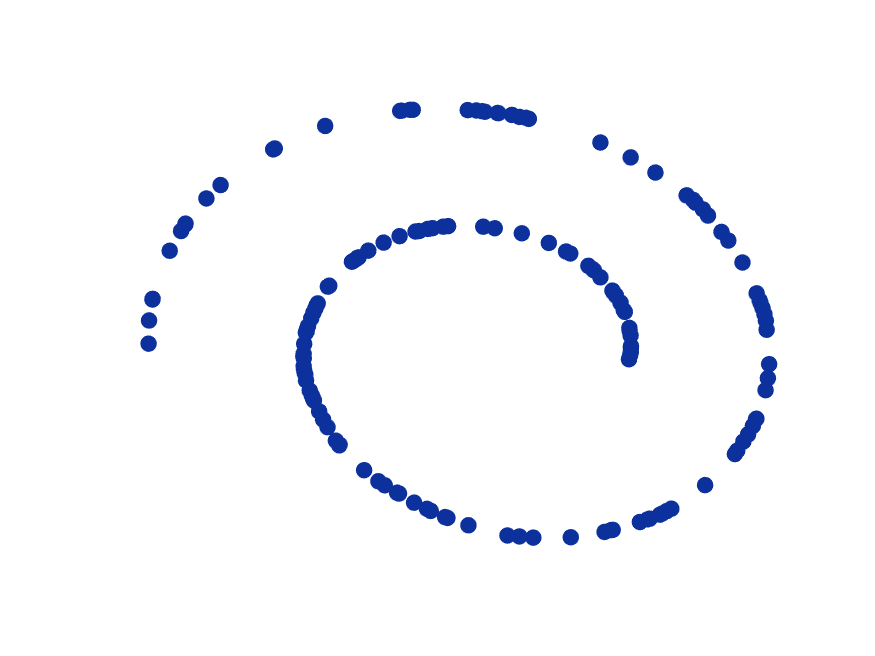}}
  \caption{}
	\label{subfig:swissvaells}
\end{subfigure}

  \caption{\label{fig:vaeCompare_swiss} Embedding of swiss roll inputs in VAE latent spaces. (a) Ground truth latent structure. (b) VAE. (c) Hyperspherical VAE. (d)  VAE with VampPrior. (e) \name.}
	
\end{figure*}

\begin{figure}[h]
\centering
\centering
\begin{subfigure}[b]{0.23\textwidth}
  \centering
	{\includegraphics[width=0.85\textwidth]{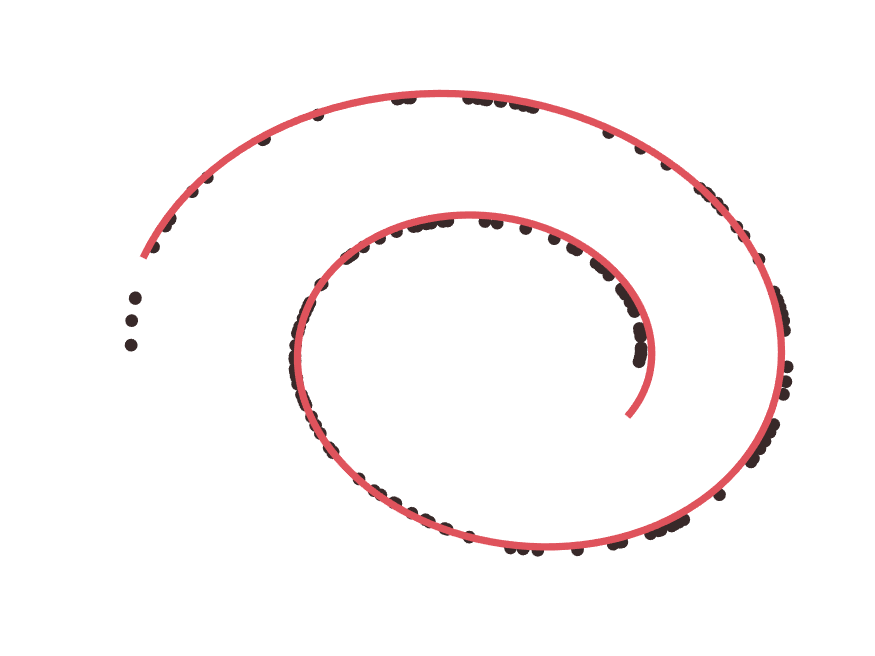}}
  \caption{}
	\label{subfig:swissTOOrbit}
\end{subfigure}
\begin{subfigure}[b]{0.23\textwidth}
  \centering
	{\includegraphics[width=0.85\textwidth]{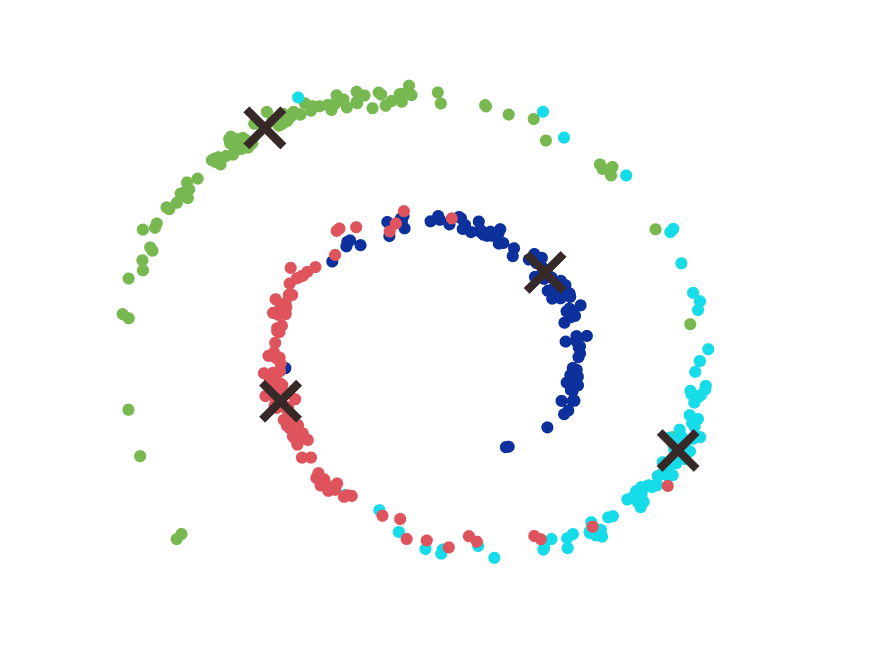}}
  \caption{}
	\label{subfig:swissAncSamp}
\end{subfigure}
\begin{subfigure}[b]{0.23\textwidth}
  \centering
	{\includegraphics[width=0.85\textwidth]{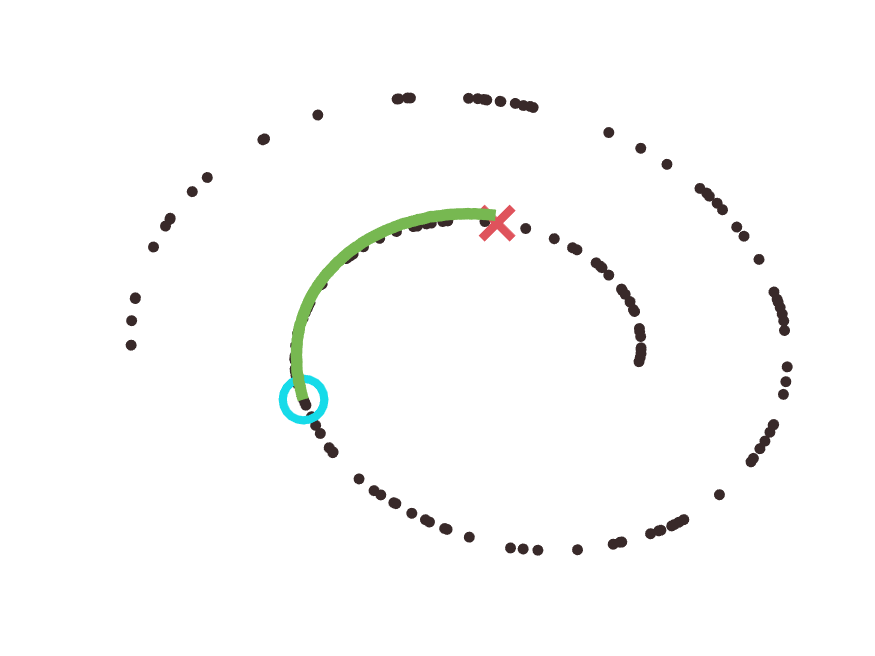}}
  \caption{}
	\label{subfig:inferPathSwiss1}
\end{subfigure}
\begin{subfigure}[b]{0.23\textwidth}
  \centering
	{\includegraphics[width=0.85\textwidth]{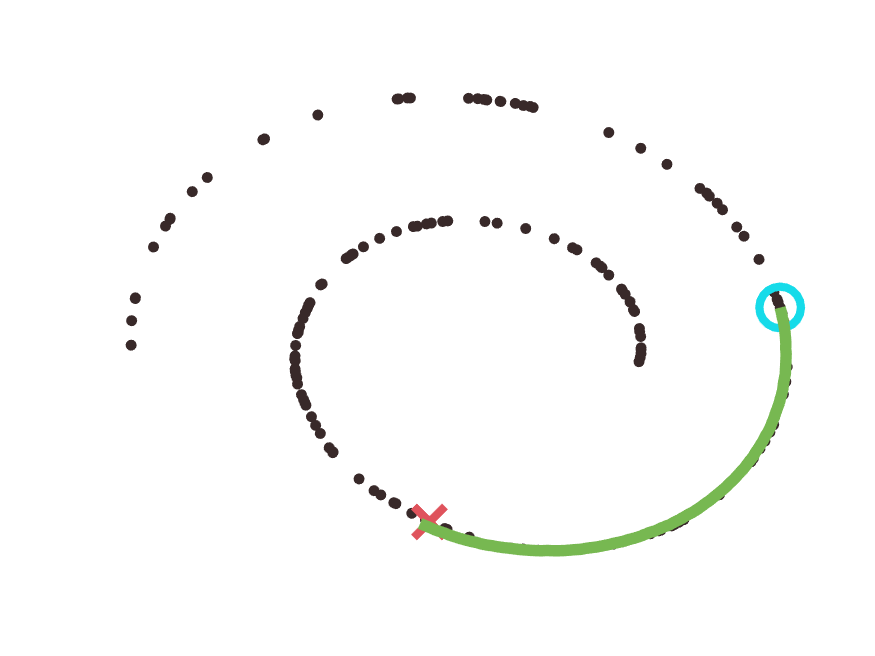}}
  \caption{}
	\label{subfig:inferPathSwiss2}
\end{subfigure}

	\caption{\label{fig:transOptOrbits_swiss} (a) The orbit of the transport operator learned on the swiss roll dataset plotted on top of encoded latent points. (b) Sampling of the latent space prior from each of the anchor points (labeled with black x's); each color indicates a separate anchor point of origin. (c-d) Transport operator paths inferred between pairs of points on the swiss roll manifold in the latent space. The cyan circle is the path starting point and the red x is the desired path ending point}

\end{figure}

\paragraph{Swiss Roll:}
We begin by applying \name{}\ to a dataset composed of 20-dimensional vector inputs that are mapped from a 2D ground truth latent space with a swiss roll structure (Fig.~\ref{subfig:swissGT}). We selected this classic manifold test structure because many VAE techniques that incorporate specific structured priors into the latent space have not demonstrated the ability to adapt to this specific geometry. The latent space is two-dimensional and the \name{} prior uses four anchor points that are spread out along the swiss roll (shown as black x's in Fig.~\ref{subfig:swissAncSamp}). Because this experiment involves only one class, the same four anchor points are used for each training sample. In the swiss roll test we learn a single operator.

Fig.~\ref{fig:vaeCompare_swiss} shows the latent space embedding for several VAE techniques. The traditional VAE with a Gaussian prior in the latent space loses the latent structure of the true data manifold because it encourages all points to cluster around the origin (Fig.~\ref{subfig:swissVAE}). The hyperspherical VAE similarly loses the true data structure because it distributes the latent points on a hypersphere (Fig.~\ref{subfig:swissHVAE}). The VAE with VampPrior is able to estimate the spiraling characteristic of the swiss roll structure (Fig.~\ref{subfig:swissVampprior}), but it is not a smooth representation of the true data manifold. By contrast, the encoded points in \name{}\ (Fig.~\ref{subfig:swissvaells}) clearly adapt to the swiss roll structure of the data.

We also utilize the swiss roll dataset to provide an intuitive understanding of how the prior in our method is formed as a combination of the learned transport operators and the encoded anchor points. Fig.~\ref{subfig:swissTOOrbit} contains the encoded latent points overlaid with the orbit of the operator learned by \name{}. Specifically, the colored line shows how the transport operator trajectory for our learned operator evolves over time when applied to a single starting point. This trajectory can be generated for any dictionary element $\mtx{\Psi}_m$: $z_t = \mathrm{expm}(\mtx{\Psi}_m \frac{t}{T})z_0$, $t = 0,...,T$. Fig.~\ref{subfig:swissAncSamp} contains latent points sampled from the prior using the sampling detailed in \eqref{eq:sampling} with increased $b$ and $\gamma$ parameters to make the sampling easier to visualize. This shows how the prior has been well-adapted to the swiss roll structure. Finally, we demonstrate how transport operators can be used to define nonlinear paths in the latent space. To generate paths between pairs of points with our learned operators, we first infer the coefficients $c^*$ between each pair. We then interpolate the path from the starting point $z_0$ as follows: $z_t = \mathrm{expm}\left(\sum_{m=1}^M{\Psi_m c^*_m t}\right) z_0$. Fig.~\ref{fig:transOptOrbits_swiss}(c-d) show two example inferred paths between points encoded on the swiss roll manifold. This experiment highlights three beneficial characteristics of \name{}\ - learning a specific latent space manifold structure, sampling points from that manifold, and generating paths directly on the manifold surface. 

\paragraph{Concentric Circle:}

\begin{figure*}[ht]

\centering
\begin{subfigure}[b]{0.19\textwidth}
  \centering
	{\includegraphics[width=0.75\textwidth]{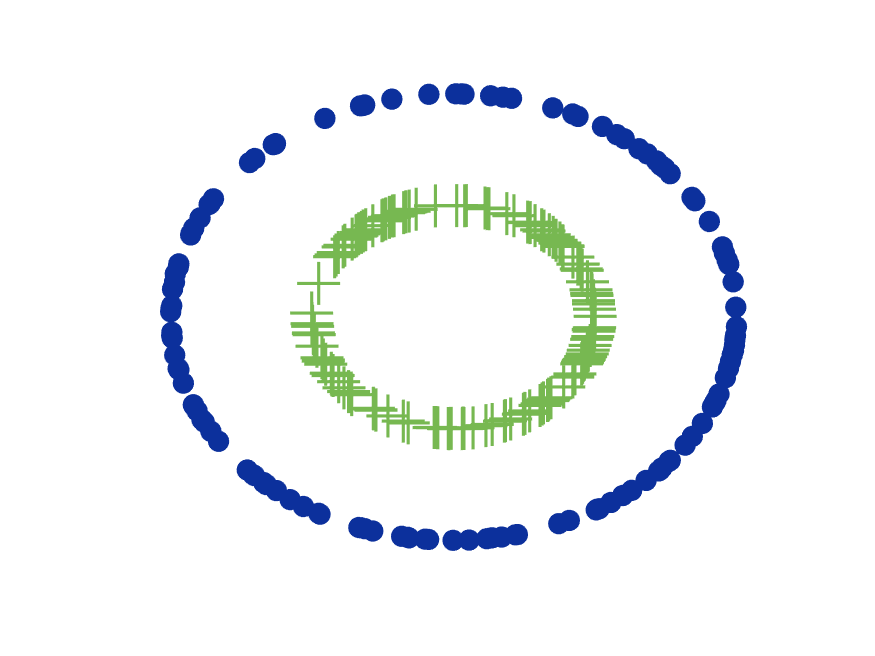}}
  \caption{}
	\label{subfig:circleGT}
\end{subfigure}
\begin{subfigure}[b]{0.19\textwidth}
  \centering
	{\includegraphics[width=0.75\textwidth]{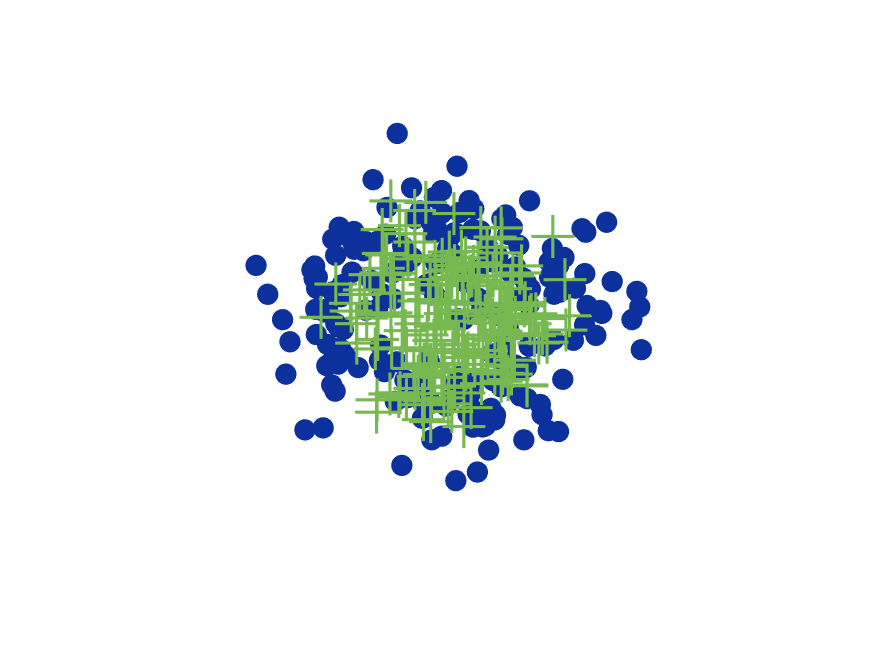}}
  \caption{}
	\label{subfig:circleVAE}
\end{subfigure}
\begin{subfigure}[b]{0.19\textwidth}
  \centering
	\includegraphics[width=0.75\textwidth]{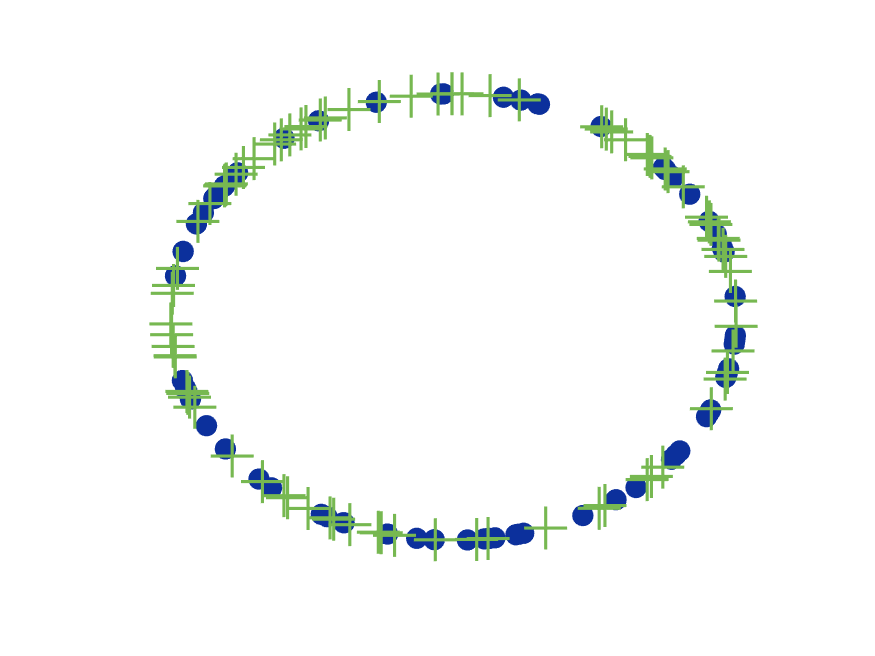}
	\caption{}
	\label{subfig:circleHVAE}
\end{subfigure}
\begin{subfigure}[b]{0.19\textwidth}
  \centering
	{\includegraphics[width=0.75\textwidth]{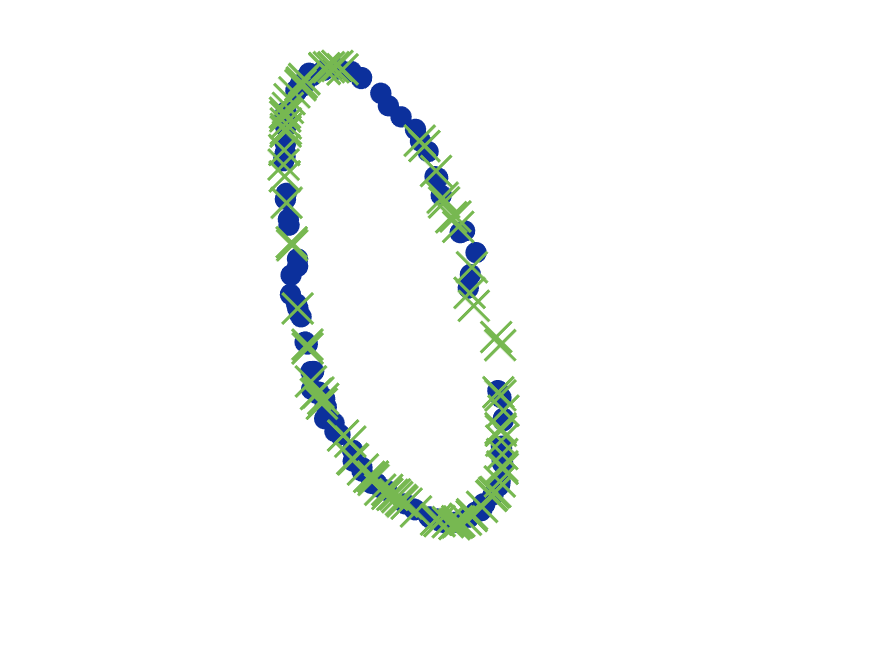}}
  \caption{}
	\label{subfig:circleVampprior}
\end{subfigure}
\begin{subfigure}[b]{0.19\textwidth}
  \centering
	{\includegraphics[width=0.75\textwidth]{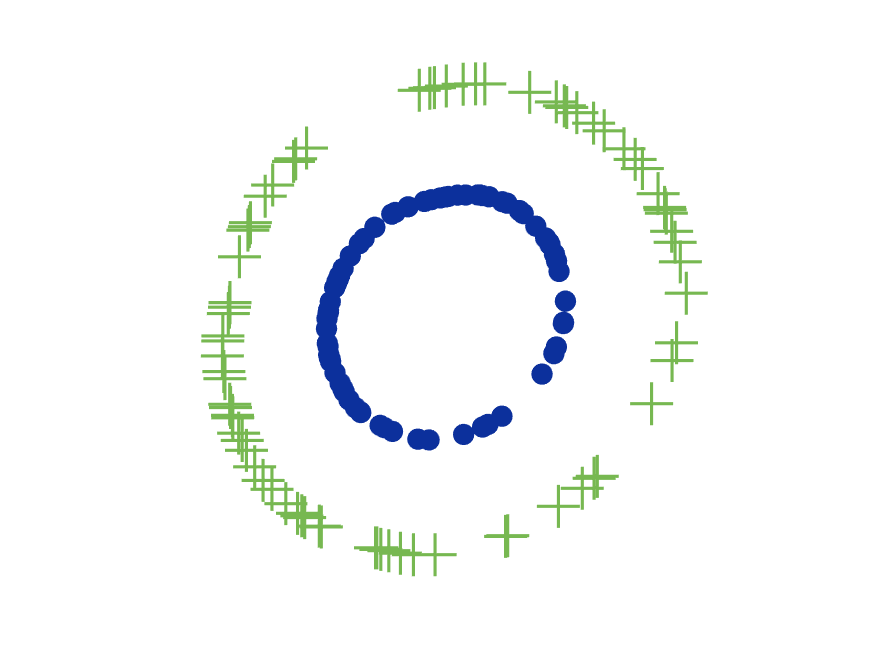}}
  \caption{}
	\label{subfig:circlevaells}
\end{subfigure}

  \caption{\label{fig:vaeCompare}  Embedding of concentric circle inputs in VAE latent spaces. (a) Ground truth latent structure. (b) VAE. (c) Hyperspherical VAE. (d)  VAE with VampPrior. (e) \name.}
	
\end{figure*}

Next we apply \name{}\ to a dataset composed of 20-dimensional data points that are mapped from a 2D ground truth latent space with two concentric circles (Fig.~\ref{subfig:circleGT}). As in the previous example, our network maps these inputs into a two-dimensional latent space. This experiment has two classes (inner circle and outer circle) so we select three anchor points per concentric circle with the anchor points evenly spaced on each circle. During training, the training points from each circle are compared against only those anchor points on the same circle manifold. In the concentric circle test we learn four operators.

This dataset in particular is well-suited for assessing how well each method is able to discriminate between the two concentric circle manifolds once points are mapped into the latent space. Fig.~\ref{fig:vaeCompare} shows the encoded latent points for several different VAE approaches. All three comparison techniques (Fig.~\ref{fig:vaeCompare}(b-d)) lose the class separation between the ground truth concentric circle manifolds. Additionally, as in  Fig.~\ref{fig:vaeCompare_swiss}, the Gaussian prior of the traditional VAE distorts the true data structure (Fig.~\ref{subfig:circleVAE}), and the VAE with VampPrior encodes a latent structure with similar characteristics to the ground truth manifold but fails to model the exact shape (Fig.~\ref{subfig:circleVampprior}). By contrast, the encoded points in the \name{}\ latent space maintain the class separation while simultaneously encoding the true circular structure. This verifies two characteristics of our approach that improve upon the traditional VAE model -- learning the prior from the data and class separation.

\paragraph{Rotated MNIST Digits:}\label{sec:rot_mnist}

The rotated MNIST dataset~\citep{lecun1998gradient} is a natural choice for demonstrating \name{}\ because it consists of real images in which we have an intuitive understanding of what the rotational transformations should look like. To define the rotated digit manifold, we specify anchor points as rotated versions of training inputs and aim to learn a transport operator that induces a latent space transformation corresponding to digit rotation. In practice this means that for each training sample we select several rotated versions of that digit as anchor points. 

\begin{figure}[h]

\centering
\begin{subfigure}[b]{0.49\textwidth}
  \centering
	{\includegraphics[width=0.98\textwidth]{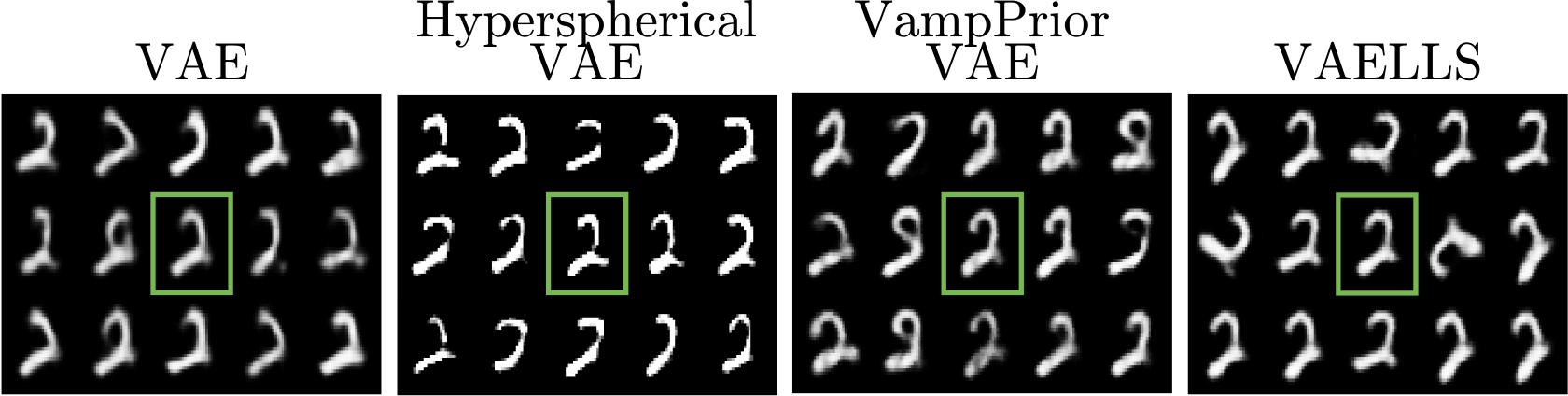}}
  \caption{}
	\label{subfig:rotDigitSample_2}
\end{subfigure}
\begin{subfigure}[b]{0.49\textwidth}
  \centering
	{\includegraphics[width=0.98\textwidth]{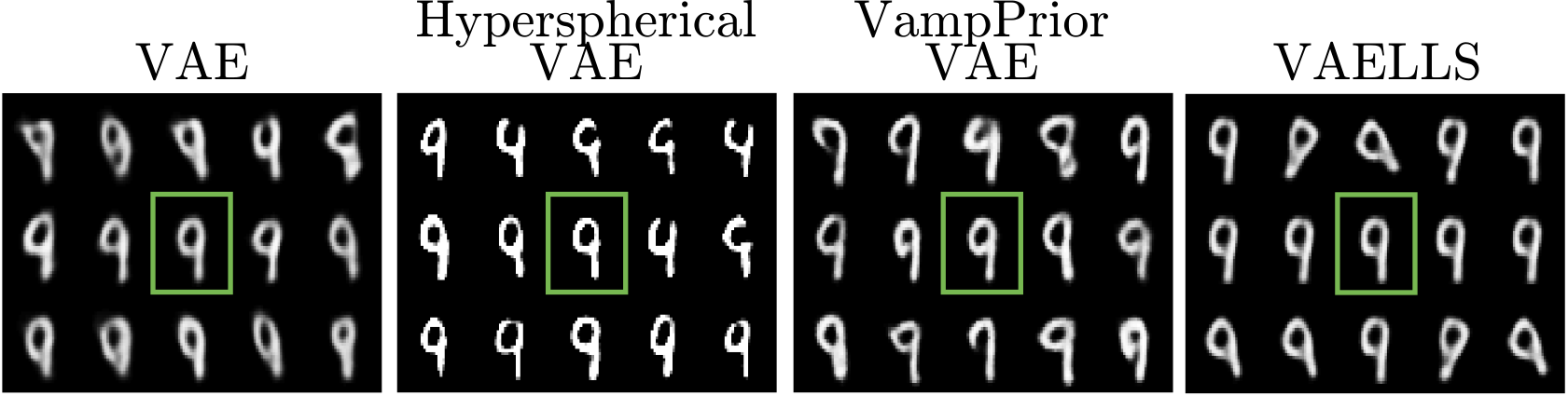}}
  \caption{}
	\label{subfig:rotDigitSample_9}
\end{subfigure}
	
  \caption{\label{fig:rotDigitSample} Examples of images decoded from latent vectors sampled from the posterior of models trained on rotated MNIST digits. In each example, the center digit (in the green box) is the decoded version of the input digit and the surrounding digits are images decoded from the sampled latent vectors. Sampling in the \name{}\ latent space results in rotations in the sampled outputs.}
	
\end{figure}

First, to highlight that we can learn a transformation model that is adapted to the rotated digit manifold, we show the result of generating data from input points in the test set using the sampling procedure described in \eqref{eq:sampling}. Fig.~\ref{fig:rotDigitSample} shows the decoded outputs of latent vectors sampled from the posterior for two example test points using four different VAE models. In each example, the center image (enclosed in a green box) is the decoded version of the input test sample. The images surrounding each center are decoded outputs of latent posterior samples. In order to visualize noticeable sampling variations, we increase the standard deviation and scale of the sampling noise in each of these models. The key result is that the \name{}\ sampling procedure using the learned transport operator leads to latent space transformations that correspond to rotations in decoded outputs. This verifies that the transport operator corresponds to movement on a learned rotated digit manifold, unlike comparison techniques which only capture natural digit variations and not specifically rotation.

\begin{figure}[h]

\centering
\begin{subfigure}[b]{0.23\textwidth}
  \centering
	{\includegraphics[width=0.98\textwidth]{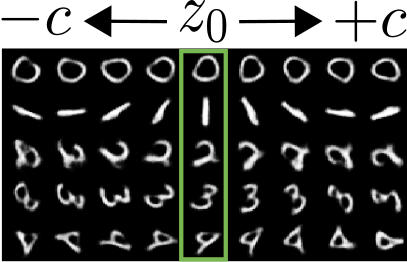}}
  \caption{}
	\label{subfig:rotDigitTransform_1}
\end{subfigure}
\begin{subfigure}[b]{0.23\textwidth}
  \centering
	{\includegraphics[width=0.98\textwidth]{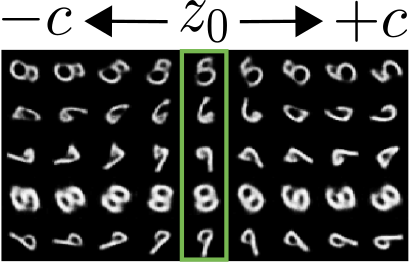}}
  \caption{}
	\label{subfig:rotDigitTransform_2}
\end{subfigure}
	
 \caption{\label{fig:rotDigitPath} Extrapolated rotation paths in the \name{}\ latent space. The center digit in each row is a decoded version of an input digit and the digits to the left and right of center show decoded outputs from latent paths extrapolated using the learned transport operator with negative and positive coefficients respectively.}
	
\end{figure}

In Fig.~\ref{fig:rotDigitPath} we show how we can extrapolate rotational paths in the latent space using our learned transport operator. To generate this figure, we randomly select example MNIST digits with zero degrees of rotation and then encode those digits to get each starting point $z_0$. The decoded versions of these initial points are shown in the middle columns (enclosed by a green box) in each figure. We then apply the learned operator with both positive and negative coefficients to $z_0$ and decode the outputs. The images to the left of center show the path generated with negative coefficients and the images to the right of center show the path generated with positive coefficients. This shows how we can generate rotated paths using the learned transport operator. It also highlights the ability for \name{}\ to define identity-preserving transformations with respect to selected input points. In these examples, the class identity of the digit is qualitatively preserved for about 180 degrees of rotation.


%

%


\paragraph{Natural MNIST Digits:}

In our final experiment, we highlight our ability to learn the natural manifold structure in MNIST digits. The anchor points are initialized by randomly selecting training examples from each digit class without any special consideration given to the specific image selection. Without a priori knowledge of the manifold structure, we have flexibility in how to parameterize our model; two parameters of specific interest are the number of transport operator dictionary elements $M$ used to define latent space transformations and the number of anchor points per class $N_a$. These parameters impact the two components of the prior definition: the learned transport operator model and the anchor points. Table~\ref{tab:evl_metric} shows how varying these parameters impacts the quantitative performance of \name{}\ as measured by estimated log-likelihood (LL) and mean-squared error (MSE) between input and reconstructed images.  Our estimated log-likelihood is based off of~\citet{burda2015importance} and the details of its computation with our model are given in Appendix~\ref{app:estLL}. The number of dictionary elements $M$ has a large impact on the estimated log-likelihood. As the number of dictionary elements increases, the value of $-\log\left[q_\phi(z \mid c,x)\right]$ decreases significantly which increases the prior term in our log-likelihood computation. This indicates that more dictionary elements enable the model to more accurately estimate paths between sampled latent points and the encoded anchor point locations. The number of anchor points does not have a clear impact on the log-likelihood and the MSE stays fairly constant as we vary the $M$ and $N_a$ parameters. For the rest of the results we use $M=8$, $N_a = 8$.

\begin{table}
\centering
\caption{Evaluation Metrics of \name{}\ Trained on MNIST}
\label{tab:evl_metric}
\begin{tabular}{|| l | l | l | l ||} 
 \hline
$M$   & $N_a$   & LL  & MSE  \\ 
 \hline
 \hline
 2 & 8 & $-4.026 \times 10^{6}$ & 0.0315\\ 
 \hline
 4 & 8 & $-1.169 \times 10^{6}$ & 0.0232 \\
 \hline
8 & 8 & $-731.91$ & 0.0244\\
 \hline
4  & 12 & $-9.248 \times 10^{5}$ & 0.0221\\ 
\hline
4 & 16 & $-7.523 \times 10^{5}$ &   0.0232\\
\hline
\end{tabular}

\end{table}

\begin{figure}[ht]

\centering
\begin{subfigure}[b]{0.49\textwidth}
 \centering
	{\includegraphics[width=0.98\textwidth]{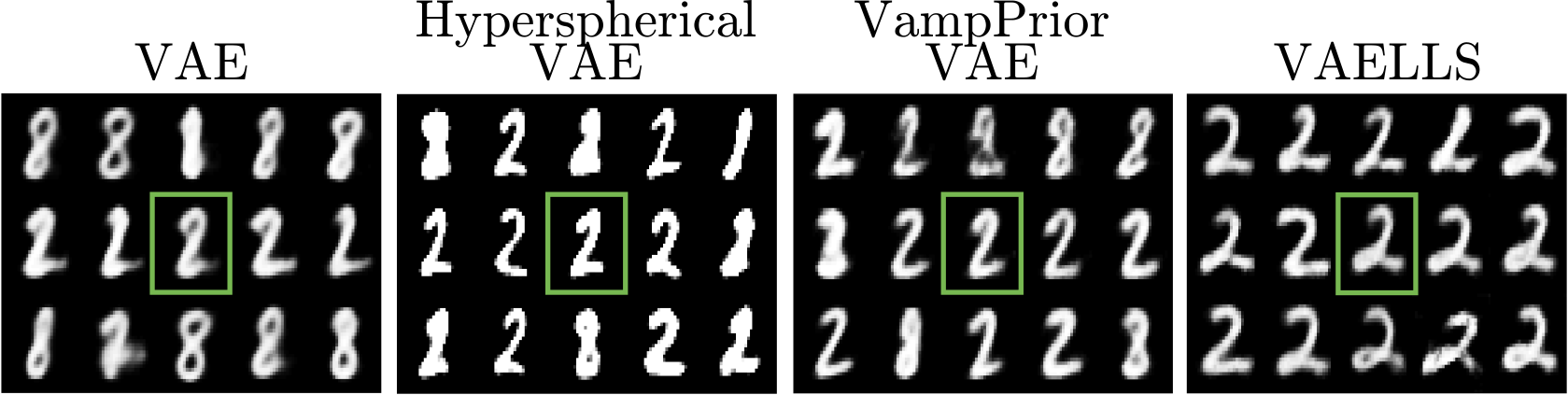}}
  \caption{}
	\label{subfig:natDigitSample_2}
\end{subfigure}
\begin{subfigure}[b]{0.49\textwidth}
  \centering
	{\includegraphics[width=0.98\textwidth]{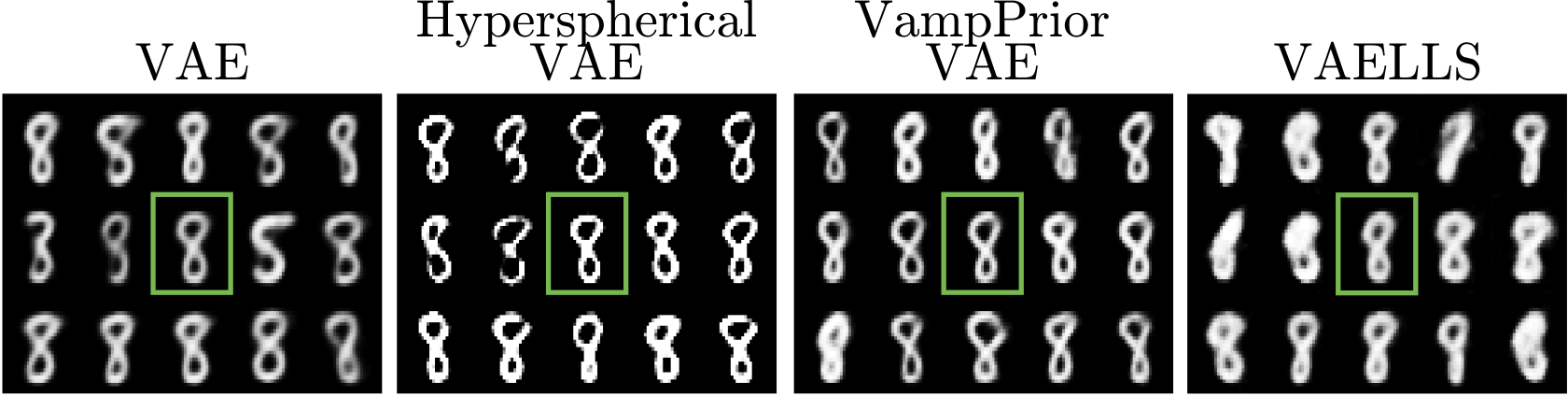}}
 \caption{}
	\label{subfig:natDigitSample_8}
\end{subfigure}
	
  \caption{\label{fig:natDigitSample} Examples of images decoded from latent vectors sampled from the posterior of a model trained on MNIST digits. In each example, the center digit (in the green box) is the encoded digit input and the surrounding digits are from the sampled latent vectors. }
\end{figure}

Fig.~\ref{fig:natDigitSample} shows the result of sampling the variational posterior in a similar manner to Fig.~\ref{fig:rotDigitSample}. Note that sampling in the \name{}\ latent space leads to natural digit transformations in the decoded outputs that maintain the class of the original digit. By contrast, the sampling in the latent space of the comparison techniques can lead to changes in class. Appendix~\ref{app:nat_mnist} contains paths generated by each of the learned transport operators that highlight how they represent natural transformation paths. This experiment demonstrates the strengths of \name{}\ in cases where the data manifold is unknown. By training on points associated with anchor points of the same class, we are able to define the prior using learned identity-preserving transformations, sample from the class manifold, and generate continuous paths on the latent space manifold.

\section{CONCLUSION}
In this paper we developed a model that has the flexibility to learn a structured VAE prior from training data by incorporating manifold transport operators into the latent space. This adaptable prior allows us to define a generative model with a latent structure that is a better fit to the data manifold. It also enables us to both interpolate and extrapolate nonlinear transformation paths in the latent space and to explicitly incorporate class separation by learning identity-preserving transformations. We verified the performance of this model on datasets with known latent structure and then extended it to real-world data to learn natural transformations. \name{}\ can be used to not only develop more realistic generative models of data but it can also be used to more effectively understand natural variations occurring in complex data. 

\subsubsection*{Acknowledgements}
We thank our anonymous reviewers for insightful comments and suggestions. This work is partially supported by NSF CAREER award CCF-1350954, ONR grant N00014-15-1-2619 and DARPA/AFRL grant FA8750-19-C-0200. This research was supported in part through research cyberinfrastructure resources and services provided by the Partnership for an Advanced Computing Environment (PACE) at the Georgia Institute of Technology, Atlanta, Georgia, USA.

\bibliography{references.bib}

\newpage
\appendix
\onecolumn

\section{DERIVATION OF THE LOSS FUNCTION}\label{app:loss_func}
In this section, we describe the details of our \name\ implementation. We define the loss function $E$ to be minimized as the approximate negative ELBO in \eqref{eq:concise_ELBO}, restated here for convenience:
\begin{gather*}
L_x(z) \equiv \log p_\theta(x\mid z) + \log p_\theta(z) - \log q_\phi(z\mid x)
\\ E(x)\equiv -\E_{u,\varepsilon}\left[L_x(T_\Psi(l(u;b))f_\phi(x)+\gamma \varepsilon)\right] + \frac{\eta}{2}\sum_{m=1}^M \lVert\Psi_m \rVert^2_F
\\ \varepsilon \sim \mathcal{N}(0,I) \quad u \sim ~\mathrm{Unif}\left(-\frac12,\frac12\right)^M.
\end{gather*}
In practice, when optimizing this loss function we approximate it with a set of $N_s$ samples:
\begin{gather}
    \widehat{E}(x) \equiv -\frac{1}{N_s}\sum_{s=1}^{N_s}L_x(T_\Psi(l(u_s;b))f_\phi(x)+\gamma\varepsilon_s)\label{app:energy}+ \frac{\eta}{2}\sum_{m=1}^M \lVert\Psi_m \rVert^2_F
    \\ \varepsilon_s \sim \mathcal{N}(0,I) \quad u_s \sim ~\mathrm{Unif}\left(-\frac12,\frac12\right)^M.\notag
\end{gather}
The deterministic mapping $l(u;b)$ is an inverse transform that maps independent uniform variates to the following factorial Laplace distribution:
\[q(c)=\prod_{m=1}^M q(c_m),\]
where for $b > 0$,
\[q(c_m)= \frac{1}{2b}\exp(- \frac{\abs{c_m}}{b}).\]
Specifically, we sample independently from each marginal $q(c_m)$ by defining the $m$th element of $l(u;b)$ as follows:
\[l_m(u;b) = -b \operatorname{sgn}(u_m)\log(1-2\abs{u_m}).\quad\]

Next we derive expressions for each expanded term in \eqref{app:energy}. For the likelihood term, we have
\begin{align*}
    -\log p_\theta(x\mid z)&=-\log\left[ (2\pi)^{\frac{-D}{2}}\sigma^{-D}\exp\left(-\frac{\norm{x-g_\theta(z)}_2^2}{2\sigma^{2}}\right)\right]
    \\ &= C_1 + \zeta_1\norm{x-g_\theta(z)}_2^2,
\end{align*}
where $\zeta_1=2^{-1}\sigma^{-2}$ is treated as a hyperparameter and $C_1 = \frac{D}{2}\log(2\pi) + D \log\sigma$. For the variational posterior term, we have
\begin{align}
    \log q_\phi(z \mid x) &= \log \int_c q_\phi(z,c \mid x) dc\label{eq:varpost_int}
    \\ &\approx \max_c \log \left[q_\phi(z \mid c,x)q(c)\right],\label{app:maxapprox}
\end{align}
where the approximation in \eqref{app:maxapprox} is motivated by the fact that the sparsity-inducing Laplace prior on $c$ typically results in joint distributions with $z$ that are tightly peaked in the coefficient space, as described in \citet{olshausen1997sparse}. Due to this tight peak in $q_\phi(z,c \mid x)$, we can approximate the integral in \eqref{eq:varpost_int} by approximating the volume under the surface in a small neighborhood around the density maximizer. Specifically, we approximate this volume by evaluating the joint density at it's maximum, weighted by the volume of a ball with a small radius centered at the maximizer. This volume approximation results in an additive constant in the log posterior that reflects the area of the chosen neighborhood, which we omit from this log-likelihood computation.  

We have
\[\begin{split}
\log \left[q_\phi(z \mid c,x)q(c)\right]=C_2-\zeta_2\norm{z-T_\Psi(c)f_\phi(x)}_2^2 - \zeta_3 \sum_{m=1}^M \abs{c_m},
\end{split}\]
where $\zeta_2=2^{-1}\gamma^{-2}$ and $\zeta_3 = b^{-1}$ are treated as a hyperparameters and $C_2 = -\frac{d}{2}\log(2\pi) - d\log\gamma+M\log \frac{1}{2b}$. Using the notation
\begin{equation}\label{eq:genCoeffInfer}
c^*(z,x;\zeta)=\argmin_c \left[ \zeta_2\norm{z-T_\Psi(c)f_\phi(x)}_2^2+ \zeta \sum_{m=1}^M \abs{c_m}\right],
\end{equation}
we have (with hyperparameter $\zeta_q$)
\begin{equation}\label{eq:post_coeffInfer}
    \log q_\phi(z \mid x) \approx C_2-\zeta_2\norm{z-T_\Psi(c^*(z,x;\zeta_q))f_\phi(x)}_2^2 - \zeta_3 \sum_{m=1}^M \abs{c^*_m(z,x;\zeta_q)}.
\end{equation}
Finally, for the prior distribution (with hyperparameter $\zeta_p$) we have
\begin{gather}
    -\log p_\theta(z)=-\log\frac{1}{N_a}\sum_{i=1}^{N_a} q_\phi(z\mid a_i)\notag
\\ \approx \log N_a-C_2-\log \sum_{i=1}^{N_a} \exp\left(-\zeta_2\norm{z-T_\Psi(c^*(z,a_i;\zeta_p))f_\phi(a_i)}_2^2-\zeta_3\sum_{m=1}^M  \abs{c^*_m(z,a_i;\zeta_p)}\right),\notag
\end{gather}
where
we use the same approximation as \eqref{app:maxapprox}.

All together, dropping additive constants and letting $z_s = T_\Psi(l(u_s))f_\phi(x)+\gamma\varepsilon_s$, we have
{\fontsize{8}{10}\[\begin{split}\widehat{E}(x)=\frac{1}{N_s}\sum_{s=1}^{N_s} \zeta_1\norm{x-g_\theta(z_s)}_2^2-\zeta_2\norm{z_s-T_\Psi(c^*(z_s,x;\zeta_q))f_\phi(x)}_2^2
- \zeta_3\sum_{m=1}^M \abs{c^*_m(z_s,x;\zeta_q)}
\\-\log\sum_{i=1}^{N_a} \exp\left(-\zeta_2\norm{z_s-T_\Psi(c^*(z_s,a_i;\zeta_p))f_\phi(a_i)}_2^2-\zeta_3\sum_{m=1}^M\abs{c^*_m(z_s,a_i;\zeta_p)}\right)+ \frac{\eta}{2}\sum_{m=1}^M \lVert\Psi_m \rVert^2_F
\end{split}\]}
In practice, one may wish to construct the prior with different constants than those used for the variational posterior term (i.e., constants $\zeta_4,\zeta_5$ instead of $\zeta_2,\zeta_3$). This substitution results in the final \name{} objective
{\fontsize{8}{10}\begin{equation}\label{eq:fullObj}\begin{split}\widehat{E}(x)=\frac{1}{N_s}\sum_{s=1}^{N_s} \zeta_1\norm{x-g_\theta(z_s)}_2^2-\zeta_2\norm{z_s-T_\Psi(c^*(z_s,x;\zeta_q))f_\phi(x)}_2^2
- \zeta_3\sum_{m=1}^M \abs{c^*_m(z_s,x;\zeta_q)}
\\-\log\sum_{i=1}^{N_a} \exp\left(-\zeta_4\norm{z_s-T_\Psi(c^*(z_s,a_i;\zeta_p))f_\phi(a_i)}_2^2-\zeta_5\sum_{m=1}^M\abs{c^*_m(z_s,a_i;\zeta_p)}\right)+ \frac{\eta}{2}\sum_{m=1}^M \lVert\Psi_m \rVert^2_F
\end{split}\end{equation}}

We allow the user to tune all of the hyperparameters during training. 

\section{DETAILS ON VAELLS TRAINING PROCEDURE}\label{app:training}

In this section we provide additional details on the general training procedure for all of the experiments. Algorithm 1 provides a detailed view of the \name{}\ training steps. For specific architecture details and parameter selections, see each experiment's respective Appendix section.

\textbf{Network training} To enhance the generative capability of the decoder, in some experiments, we use a warm-up as in~\citet{tomczak2018vae}. Our warm-up includes updates to the network weights driven by only the reconstruction loss. During warm up there is no Gaussian sampling in the latent space and the sampled Laplace distribution for the transport operator coefficients has a large $\zeta_3$ parameter which encourages sampling coefficients that are very close to zero . As mentioned in Section~\ref{sec:experiments}, during~\name{}\ training we alternate between steps where we update the network weights and anchor points while keeping the transport operators fixed and steps where we update the transport operators while keeping the network weights and anchor points fixed. As we alternate between these steps, we vary the weights on the objective function terms. Specifically, we decrease the importance of the prior terms during the steps updating the network weights and decrease the importance of the reconstruction term during the steps updating the transport operators.

\textbf{Transport operator learning} As mentioned in Section~\ref{sec:experiments}, one component of computing the prior and posterior objective is the coefficient inference between the sampled point $z$ and the neural network encoding $f_{\phi}(x)$ as well as all the encoded anchor points $f_{\phi}(a_i)$. Note that the transport operator objective is non-convex which may result in coefficient inference optimization arriving at a poor local minima. This issue can be avoided by performing the coefficient inference between the same point pair several times with different random initializations of the coefficients and selecting the inferred coefficients that result in the lowest final objective function. We leave the number of random initializations as a parameter to select during training.

Transport operator coefficient inference is best performed when the magnitude of the latent vector entries is close to the range $[-1,1]$. Because of this, we allow for the selection of a scale factor that scales the latent vectors prior to performing coefficient inference. In practice, we inspect the magnitude of the latent vectors after warm-up training steps and select a scale factor that adapts the largest magnitudes of the latent vector entries to around 1.

During every transport operator update step, our optimization routine checks whether the transport operator update improves the portion of the objective that explicitly incorporates the transport operator:
\begin{equation}
    \begin{split}\widehat{E}_{\mathrm{transopt}}(x)=\frac{1}{N_s}\sum_{s=1}^{N_s} -\zeta_2\norm{z_s-T_\Psi(c^*(z_s,x;\zeta_q))f_\phi(x)}_2^2
- \zeta_3\sum_{m=1}^M \abs{c^*_m(z_s,x;\zeta_q)}
\\-\log\sum_{i=1}^{N_a} \exp\left(-\zeta_4\norm{z_s-T_\Psi(c^*(z_s,a_i;\zeta_p))f_\phi(a_i)}_2^2-\zeta_5\sum_{m=1}^M\abs{c^*_m(z_s,a_i;\zeta_p)}\right)
\end{split}
\end{equation}
 If this portion of the objective does not improve with a gradient step on the dictionary then we reject this step and decrease the transport operator learning rate. If the transport operator portion of the objective does improve with the gradient step on the dictionary then we accept this step and increase the transport operator learning rate. This helps us settle on an appropriate learning rate and prevents us from making ineffective updates to the dictionaries. We also set a maximum transport operator learning rate which varies based on the experiment. 

Another unique consideration during transport operator training is that we generally assume that the transport operator training points are close on the manifold. In the formulation of the variational posterior, this is a reasonable assumption because $z$ is a sample originating from $f_{\phi}(x)$. However, in the prior formulation, while the anchor points are generally sampled in a way that encourages them to be evenly spaced in the data space, it is unlikely that every latent vector associated with a data point is close to every anchor point. In order to aid in constraining training to points that are relatively close on the manifold, we provide the training option of defining the prior with respect to only the anchor point closest to $z$ rather than summing over all the anchor points:

\begin{equation}
p_\theta(z)=q_\phi(z\mid a^*), \label{eq:prior_singlealpha}
\end{equation}
 where $a^*$ is the anchor that is estimated to be closest to $z$. Since we do not have ground truth knowledge of which anchor point is closest to a given training point on the data manifold, we estimate this by inferring the coefficients that represent the estimated path between $z$ and every $a_i$. We then select $a^*$ as the anchor point with the lowest objective function (i.e., $\zeta_4\norm{z_s-T_\Psi(c^*(z_s,a_i;\zeta_p))f_\phi(a_i)}_2^2+\zeta_5\sum_{m=1}^M\abs{c^*_m(z_s,a_i;\zeta_p)}$) after coefficient inference. This objective function defines how well $a_i$ can be transformed to $z_s$ using the current transport operator dictionary elements $\Psi$.

There are several hyperparameters that need to be tuned in this model. However, we have found through experimentation that the model is robust to changes in several of the parameters (such as $\zeta_2$, $\zeta_3$,$\zeta_4$). The hyperparameters that were shown to have the largest effect on training effectiveness were:
\begin{itemize}
    \item The weight on the reconstruction term ($\zeta_1$) - use this in combination with warm-up steps to ensure reasonable reconstruction accuracy from the decoder.
    \item The posterior coefficient inference weight ($\zeta_q$)- this is the weight on the sparsity regularizer term used in the objective (\ref{eq:genCoeffInfer}) during coefficient inference between points in the \emph{posterior} term. If this weight is too large, then inference can result in zero coefficients for all the operators which is not informative.
    \item The prior coefficient inference weight ($\zeta_p$) - this is the weight on the sparsity regularizer term used during coefficient inference between points in the \emph{prior} term.
    \item Number of restarts used during coefficient inference for transport operator training.
    \item Starting $lr_{\psi}$ - As mentioned above, we do vary $lr_{\psi}$ during transport operator training depending on whether our training steps are successful or not. If this learning rate starts too high, it can result in many unsuccessful steps with no updates on the transport operator dictionaries which greatly slows down training.
\end{itemize}

\begin{algorithm}[H]\label{alg:training}
\SetAlgoLined
 \KwData{Training samples $\mathcal{X}$, anchor points $\{a_1,...,a_{N_a}\}$ selected from the input space}
 \KwResult{Trained transport operator dictionary elements $\{\mtx{\Psi}_1,...\mtx{\Psi}_M\}$, network weights $\mtx{\phi}$ and $\mtx{\theta}$, fine-tuned anchor points $\{a_1,...,a_{N_a}\}$}
 $\mtx{\Psi}$, $\mtx{\phi}$, $\mtx{\theta} \leftarrow$ randomly initialize dictionaries, network weights\;
 \For{$k = 0,....,N$}{
  Sample mini-batch from $\mathcal{X}$: $\{x_1,...,x_{N_s}\}$\;
  
  \For{$s = 0,....,N_s$}{
    Encode samples: $\mu_s \leftarrow f_{\phi}(x_s)$\;
    Sample $u_s \sim ~\mathrm{Unif}\left(-\frac12,\frac12\right)^M$\;
   $\widehat{c}_s \leftarrow l(u_s)$\;
    Sample $z_s$ from $q_\phi(z_s \mid \widehat{c}_s ,x_s) \sim \mtx{T}_\Psi(\widehat{c}_s )\mu_s + \gamma \varepsilon_s$\;
    Decode sampled vectors: $\widehat{x}_s \leftarrow g_\theta(z_s)$\;
    $c^*(z_s,x_s) \leftarrow $ Infer\_Coefficients$(z_s,x_s,d_{\text{start}},d_{\text{stop}})$\;
    \For{$i = 0,....,N_a$}{
        \For{$r = 0,....,\mathrm{num\_restart}$}{
            $c^{(r)}(z_s,a_i) \leftarrow $ Infer\_Coefficients$(z_s,a_i,d_{\text{start}},d_{\text{stop}})$\;

            $E_{c}(c^{(r)}(z_s,a_i)) \leftarrow \log \left[q_\phi(z_s \mid c^{(r)}(z_s,a_i),a_i)q(c^{(r)}(z_s,a_i))\right] $
        }
        $c^*(z_s,a_i) \leftarrow \argmax_r E_{c}(c^{(r)}(z_s,a_i))$
    }
  }
  Calculate $\widehat{E}$ in \eqref{eq:fullObj} using $\widehat{x}_s$, $c^*(z_s,x_s)$, and $c^*(z_s,a_i)$ for $s = 0,....,N_s$ and $i = 0,....,N_a$\;
  $\mtx{\Psi}_{\text{new}} \leftarrow \mtx{\Psi} - lr_{\Psi}\frac{\delta\widehat{E}}{\delta \Psi}$\;
        \eIf{$\widehat{E}_{\mathrm{transopt}}(\mtx{\Psi}_{\mathrm{new}}) < \widehat{E}_{\mathrm{transopt}}(\mtx{\Psi})$}{
        $\mtx{\Psi} \leftarrow \mtx{\Psi}_{\text{new}}$\;
        $lr_{\Psi} \leftarrow \frac{lr_{\Psi}}{\text{decay}}$
        }
        {$lr_{\Psi} \leftarrow lr_{\Psi}\cdot\text{decay}$}
    $\mtx{\phi} \leftarrow \mtx{\phi} - lr_{\text{net}}\frac{\delta\widehat{E}}{\delta \phi}$\;
    $\mtx{\theta} \leftarrow \mtx{\theta} - lr_{\text{net}}\frac{\delta\widehat{E}}{\delta \theta}$\;
    $\mtx{a} \leftarrow \mtx{a} - lr_{\text{anchor}}\frac{\delta\widehat{E}}{\delta a}$\;
 }
 \caption{Training of network weights, transport operators, and anchor points}
\end{algorithm}

\begin{algorithm}
\KwData{Latent vector z, input data x}
 \KwResult{Inferred coefficient vector $c$ }
Initialize $c$: $c_m \sim \mathrm{Unif}[d_{\text{start}},d_{\text{stop}}]$\;
  Fix $c$ to $c^* \leftarrow  \argmax_c\log \left[q_\phi(z \mid c,x)q(c)\right]$\;
\caption{Infer\_Coefficients$(z,x,d_{\text{start}},d_{\text{stop}})$}
\end{algorithm}
\textbf{Comparison techniques} We implemented the hyperspherical VAE~\citep{davidson2018hyperspherical} using the code provided by the authors: \url{https://github.com/nicola-decao/s-vae-pytorch}. For the concentric circle and swiss roll experiments we used the network specified in Table~\ref{tab:swissNet}. For MNIST experiments, we used the network architecture from their MNIST experiments, and we dynamically binarized the MNIST inputs as they did. We implemented the VAE with VampPrior~\citep{tomczak2018vae} model using the code provided by the authors: \url{https://github.com/jmtomczak/vae_vampprior}. For the concentric circle and swiss roll experiments, we used the network architecture specified in Table~\ref{tab:swissNet} and 100 pseudoinputs for each experiment. For the MNIST experiments we adapted the network in Table~\ref{tab:rotMNISTNet} to add a linear layer between the final convTranspose layer and the sigmoid layer. We used 500 pseudoinputs in each of the MNIST experiments. The VAE with VampPrior MNIST tests were also performed on dynamically binarized MNIST data. We implemented our own VAE code with the same network architectures detailed in Tables~\ref{tab:swissNet} and \ref{tab:rotMNISTNet}.

\section{VARIATIONAL POSTERIOR CONTOURS}
\begin{figure*}[ht]

\centering
\begin{subfigure}[b]{0.3\textwidth}
  \centering
	{\includegraphics[width=0.98\textwidth]{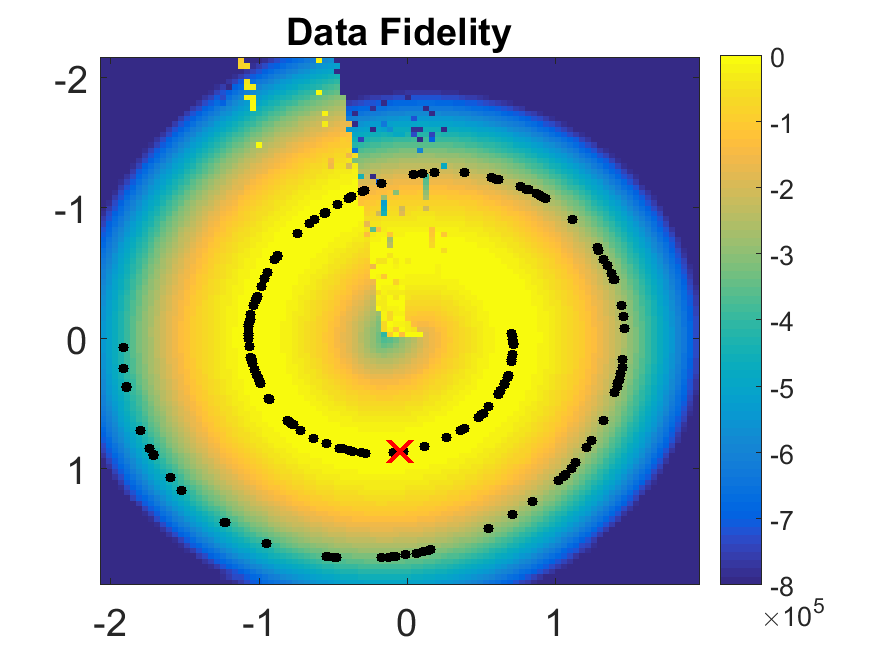}}
  \caption{}
	\label{subfig:contourFid}
\end{subfigure}
\begin{subfigure}[b]{0.3\textwidth}
  \centering
	{\includegraphics[width=0.98\textwidth]{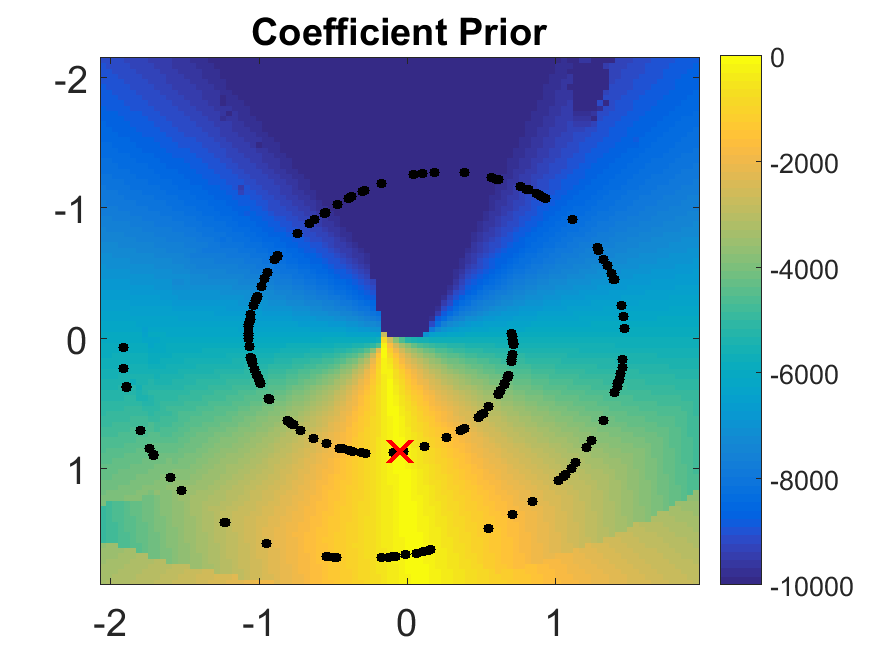}}
  \caption{}
	\label{subfig:contourSparse}
\end{subfigure}
\begin{subfigure}[b]{0.3\textwidth}
  \centering
	\includegraphics[width=0.98\textwidth]{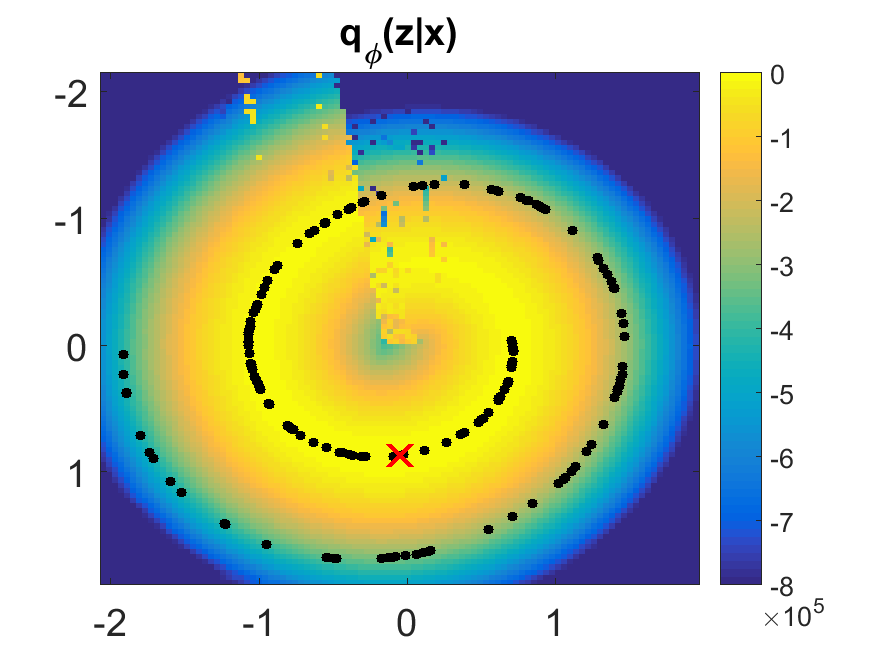}
	\caption{}
	\label{subfig:contourVarPost}
\end{subfigure}

  \caption{\label{fig:swissContour} Variational posterior contour plots. The red x is the location of the encoded point $z$. The black dots are encoded data points. (a) Contour of data fidelity term (b) Contour of coefficient prior term (c) Contour of $\log q_{\phi}(z\mid x)$ }
	
\end{figure*}

In order to more intuitively visualize the learned variational posterior we show contour plots of the variational posterior given an input point. The variational posterior (\ref{eq:post_coeffInfer}) consists of two terms: the data fidelity term $\left(-\zeta_2\norm{z-T_\Psi(c^*(z,x;\zeta_q))f_\phi(x)}_2^2\right)$ and the coefficient prior term $\left(-\zeta_3 \sum_{m=1}^M\abs{c^*_m(z,x;\zeta_q)}\right)$. The data fidelity term expresses the probability of $z$ given an input point $x$. Fig.~\ref{subfig:contourFid} shows a contour of the data fidelity term. The red x is the location of the encoded point and the black dots are encoded data samples. This shows how the data fidelity term maps out the contour of the latent swiss roll manifold around the encoded point. The coefficient prior is a Laplace distribution which encourages the coefficients to be tightly peaked around zero. Fig.~\ref{subfig:contourSparse} shows the contour of the prior term. There is a high log probability in the slice of the space that can be reached by applying small coefficients to the transport operators and the log probability reduces in parts of the space that require larger coefficient values to reach.

The final variational posterior in Fig.~\ref{subfig:contourVarPost} is a combination of these two terms. In our experimental setting, the data fidelity term is the dominant term that characterizes the variational posterior. However, as the weights $\zeta_2$ and $\zeta_3$ vary, the contribution of each term towards the variational posterior will change.

\section{ESTIMATED LOG-LIKELIHOOD}\label{app:estLL}

We estimate the log-likelihood using the importance sampling from~\cite{burda2015importance} with 500 datapoints and 100 samples per datapoint. We follow the derivation from Appendix~\ref{app:loss_func} for computing $\log p_{\theta}(x\mid z)$, $\log q_{\phi}(z\mid x)$, and $\log p_{\theta}(z)$ and include all the constants defined there. We set $\zeta_1$ to the same value used during training. While we hand-select $\zeta_2$, $\zeta_3$, $\zeta_4$, and $\zeta_5$ during training, when computing the estimated log-likelihood, we compute each of these values based on the parameters $\gamma$ and $b$ which are used for sampling the latent vectors as specified in \eqref{eq:sampling}. To be explicit,  $\zeta_2 = \zeta_4 = 2^{-1}\gamma^{-2}$ and $\zeta_3 = \zeta_5 = b^{-1}$. When we infer coefficeints as in \eqref{eq:genCoeffInfer}, we set $\zeta_2 = 1$ and we set the hyperparameters $\zeta_q$ and $\zeta_p$ (which are used for the posterior and prior respectively) to the same values used during training which yield successful coefficient inference. As noted by \eqref{app:maxapprox}, the approximation of $\log \int_c q_\phi(z,c \mid x) dc$ will result in an additive constant that we do not include in our estimated log-likelihood computation.

The results presented in Table~\ref{tab:evl_metric} show a wide variation in estimated log-likelihood over the trials. These values are largely dominated by the data fidelity term in the prior which decreases significantly as the number of dictionaries increase from 2 to 8. Because this experiment sets $\gamma = 0.001$, any variation in the data fidelity terms for the $\log q_{\phi}(z\mid x)$ and $\log p_{\theta}(z)$ are magnified when they are multiplied by $\zeta_2 = \zeta_4 = \frac{1}{2 \left(0.001^2\right)}$. 

\section{SWISS ROLL EXPERIMENT}\label{app:swiss_roll}

Tables~\ref{tab:swissNet} and~\ref{tab:swissParams} contain the VAELLS network architecture and parameters for the swiss roll experiment. In this experiment, we sample our ground truth 2D data manifold from a swiss roll and then map it to the 20-dimensional input space using a random linear mapping. We use 1000 swiss roll training points and randomly sample swiss roll test points. We initialize anchor points as points that are spaced out around the swiss roll prior to mapping to the 20-dimensional input space. We allow for the anchor points to be updated; however, these updates result in negligible changes to the anchor points. As described in Section~\ref{app:training}, in this experiment we define the prior only with respect to the anchor points that are estimated to be closest to each training point.

\begin{table}[t]
\centering
\caption{Network Architecture for Swiss Roll and Concentric Circle Experiments}
\label{tab:swissNet}
\begin{tabular}{||l l||} 
 \hline
 Encoder Network & Decoder Network  \\ 
 \hline
 Input $\in \mathbb{R}^{20}$ & Input $\in \mathbb{R}^2$  \\ 
 Linear: 512 Units & Linear: 512 Units  \\
 ReLU & ReLU \\
 Linear: 2 Units & Linear: 20 Units  \\
 \hline
\end{tabular}

\end{table}

\begin{table}[t]
\centering
\caption{Training Parameters for Swiss Roll Experiment}
\label{tab:swissParams}
\begin{tabular}{||l||} 
 \hline
 VAELLS Training - Swiss Roll \\ 
 \hline
 batch size: 30   \\ 
 training steps: 3000   \\
 latent space dimension ($z_{dim}$): 2 \\
 $N_s: 1$ \\
 $lr_{\mathrm{net}}: 10^{-4}$  \\
 $lr_{\mathrm{anchor}}: 10^{-4}$ \\
 starting $lr_{\Psi}: 5 \times 10^{-5}$  \\ 
 max $lr_{\Psi}: 0.05$ \\
 $\zeta_1:$ 0.01   \\
 $\zeta_2:$ 1   \\
 $\zeta_3:$ 1   \\
 $\zeta_4:$ 1   \\
 $\zeta_5:$ 0.01  \\
 $\zeta_q$: $1 \times 10^{-6}$\\
 $\zeta_p$:  $5 \times 10^{-5}$\\
 $\eta:$ 0.01   \\
 number of network and anchor update steps: 20  \\
 weight on prior terms during net update steps: 0.01 \\
 number of $\Psi$ update steps: 20 \\
 weight on recon term during net update steps: 0.001 \\
 $\gamma_{\mathrm{post}}:$ 0.001\\
 warm-up steps: 0\\
 number of restarts for coefficient inference: 2 \\
 $M:$ 1\\
 number of anchors: 4\\
 latent space scale: 1 \\
 define prior with respect to closest anchor point: yes \\
 \hline
\end{tabular}

\end{table}

\section{ANALYSIS OF SENSITIVITY TO ANCHOR POINTS}\label{app:anchor}

The anchor points are a key feature needed to specify the learned latent prior and its important to understand the role that anchor point selection plays in the success of learning the data manifold. We investigate this question on the swiss roll manifold with a known latent structure that will allow us to determine the success of the \name{} training procedure as we vary the anchor points.

\begin{figure*}[h]

\centering
\begin{subfigure}[b]{0.2\textwidth}
  \centering
	{\includegraphics[width=0.95\textwidth]{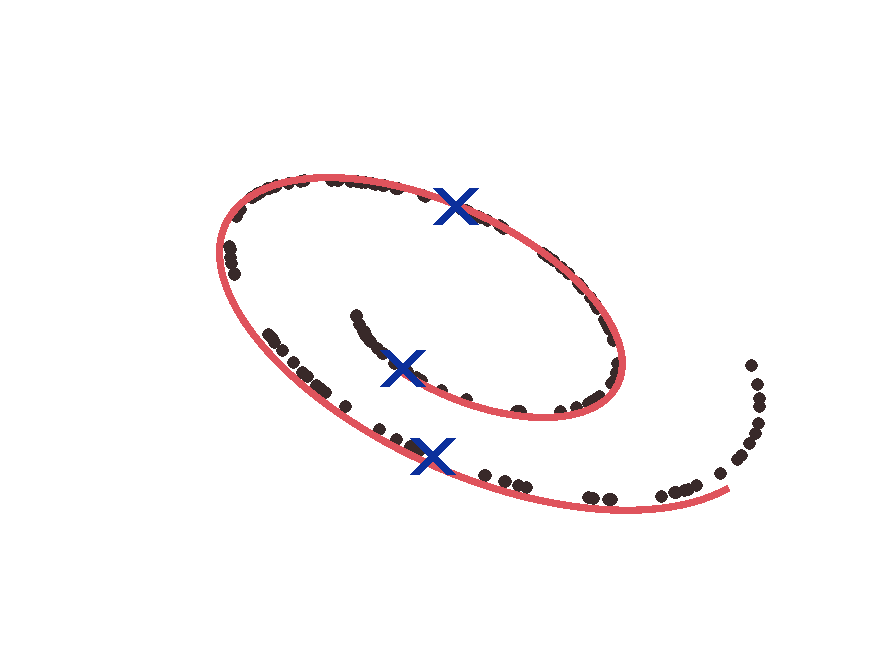}}
  \caption{}
	\label{subfig:anchorVary_3}
\end{subfigure}
\begin{subfigure}[b]{0.2\textwidth}
  \centering
	{\includegraphics[width=0.95\textwidth]{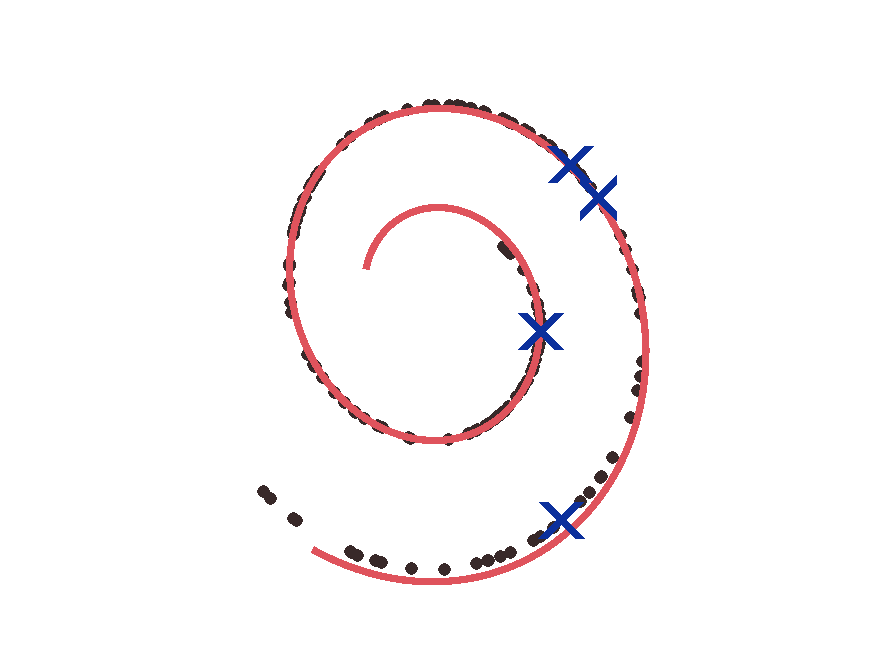}}
  \caption{}
	\label{subfig:anchorVary_4}
\end{subfigure}
\begin{subfigure}[b]{0.2\textwidth}
  \centering
	\includegraphics[width=0.95\textwidth]{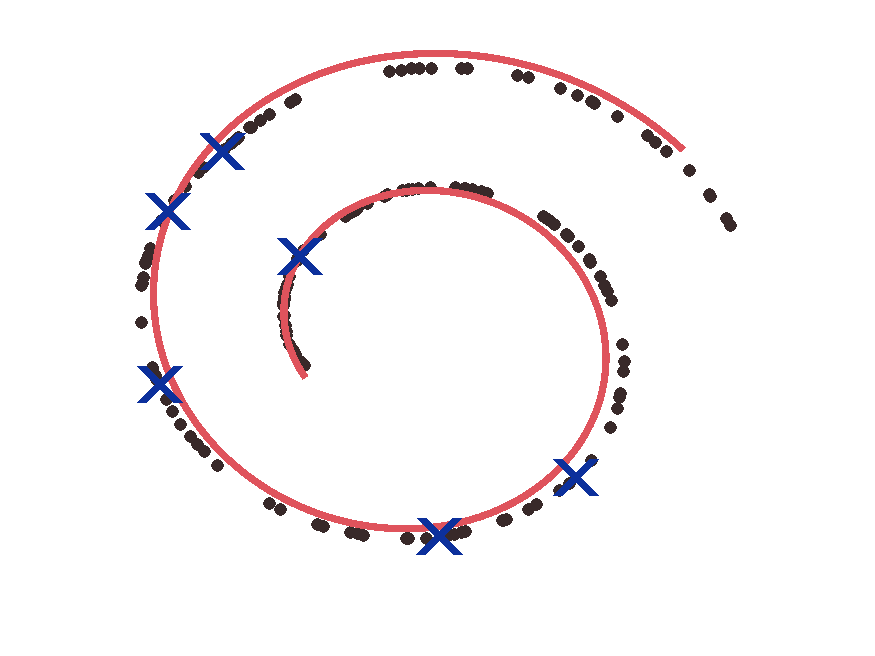}
	\caption{}
	\label{subfig:anchorVary_6}
\end{subfigure}
\begin{subfigure}[b]{0.2\textwidth}
  \centering
	{\includegraphics[width=0.95\textwidth]{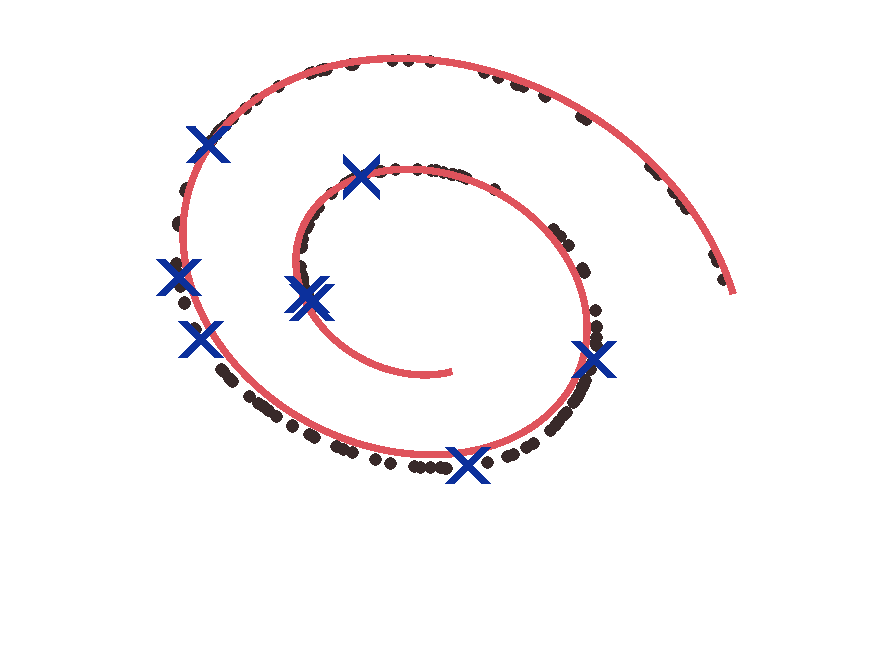}}
  \caption{}
	\label{subfig:anchorVary_8}
\end{subfigure}

  \caption{\label{fig:anchorVary} Outputs of trials where \name{} learns the correct swiss roll manifold while the number of anchor points is varied and the anchor point locations are randomly selected on the manifold. The black dots are encoded data points, the blue x's are the encoded anchor point locations, and the red line is the generated orbit of the learned transport operator. Each plot shows a result with a different number of anchor points: (a) 3 anchor points (b) 4 anchor points (c) 6 anchor points (d) 8 anchor points.}
	
\end{figure*}

\begin{figure*}[h]

\centering
\begin{subfigure}[b]{0.2\textwidth}
  \centering
	{\includegraphics[width=0.95\textwidth]{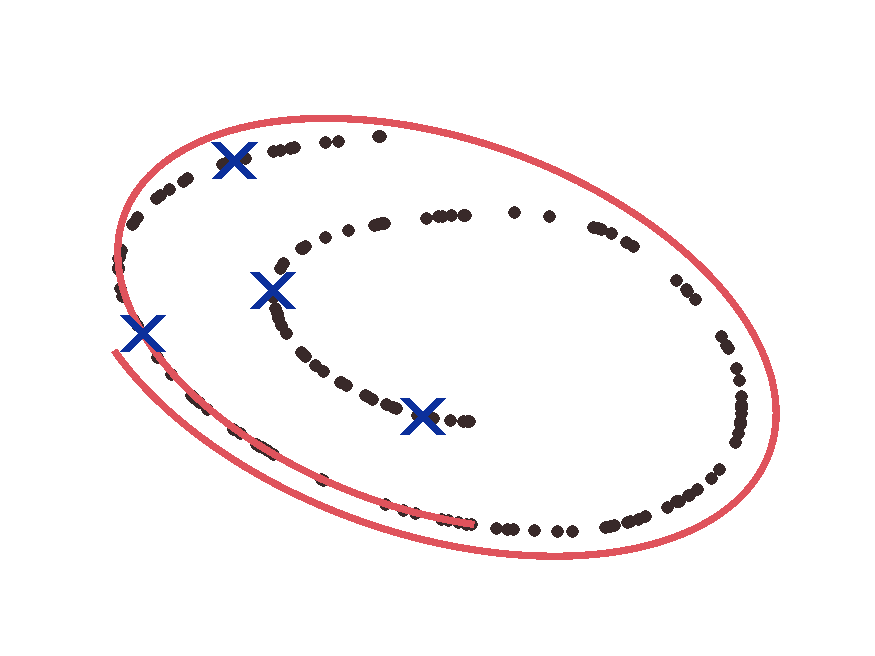}}
  \caption{}
	\label{subfig:anchorFail_4}
\end{subfigure}
\begin{subfigure}[b]{0.2\textwidth}
  \centering
	{\includegraphics[width=0.95\textwidth]{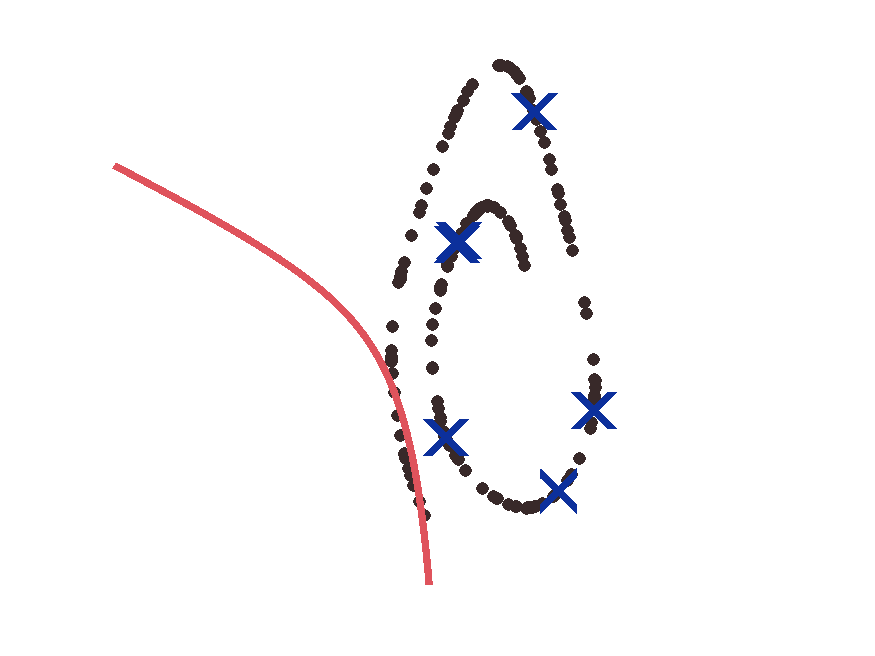}}
  \caption{}
	\label{subfig:anchorFail_6}
\end{subfigure}
\begin{subfigure}[b]{0.2\textwidth}
  \centering
	\includegraphics[width=0.95\textwidth]{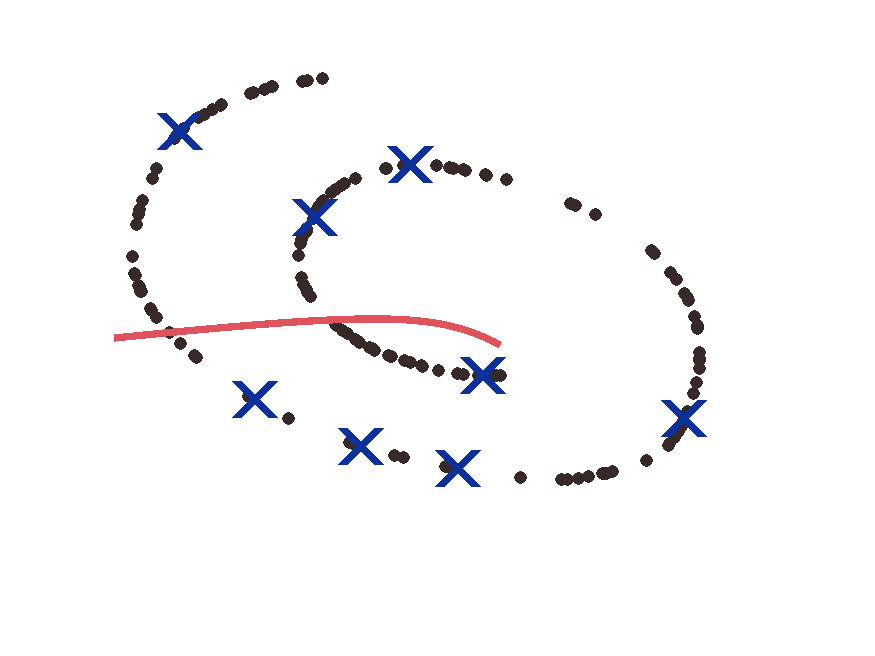}
	\caption{}
	\label{subfig:anchorFail_8}
\end{subfigure}

  \caption{\label{fig:anchorFail} Outputs of trials where \name{} fails to learn the correct swiss roll manifold while the number of anchor points is varied and the anchor point locations are randomly selected on the manifold. The black dots are encoded data points, the blue x's are the encoded anchor point locations, and the red line is the generated orbit of the learned transport operator. Each plot shows a result with a different number of anchor points: (a) 4 anchor points (b) 6 anchor points (c) 8 anchor points.}
	
\end{figure*}

We begin by varying the number and location of anchor points. In the initial formulation of the swiss roll experiment, the anchor points are selected to be well distributed around the swiss roll manifold. Fig.~\ref{subfig:swissAncSamp} shows the locations of the anchor points used for experiment detailed in the paper. The anchor points are well spaced around the swiss roll manifold.  Fig.~\ref{subfig:swissTOOrbit} shows the learned operator from this experiment which successfully generates paths with the desired swiss roll manifold structure. 
While the experiment we presented uses four well-spaced anchor points, we can vary both the number and locations of the anchor points and still learn a mapping to a swiss roll latent structure and a transport operator that generates paths along the swiss roll. Fig.~\ref{fig:anchorVary} shows examples of tests where \name{} successfully learns the swiss roll structure in the latent space with different numbers of anchor points. While we achieve success with randomly positioned anchor points, there are tests where the anchor points are relatively evenly spaced over the manifold and the transport operators do not learn the swiss roll structure. Fig.~\ref{fig:anchorFail} shows examples of tests where \name{} fails to learn the swiss roll manifold even when the anchor points are relatively evenly spaced on the encoded manifold. These results indicate that the success of learning transport operators to represent the true latent manifold depends on a factor other than the anchor point locations. 

\begin{figure*}[h]

\centering
\begin{subfigure}[b]{0.17\textwidth}
  \centering
	{\includegraphics[width=0.95\textwidth]{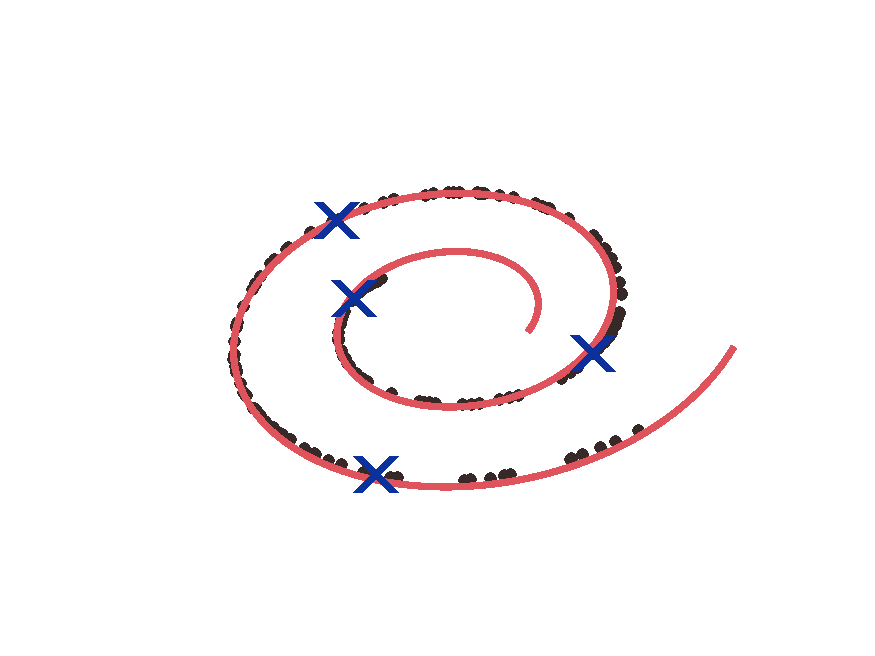}}
  \caption{}
	\label{subfig:anchorInit_success_1}
\end{subfigure}
\begin{subfigure}[b]{0.17\textwidth}
  \centering
	{\includegraphics[width=0.95\textwidth]{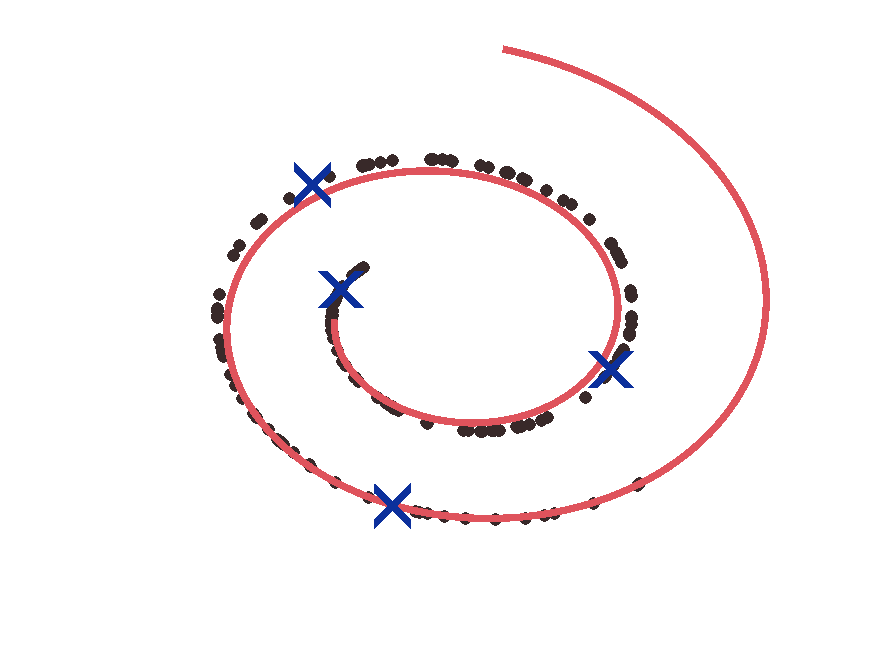}}
  \caption{}
	\label{subfig:anchorInit_success_2}
\end{subfigure}
\begin{subfigure}[b]{0.17\textwidth}
  \centering
	\includegraphics[width=0.95\textwidth]{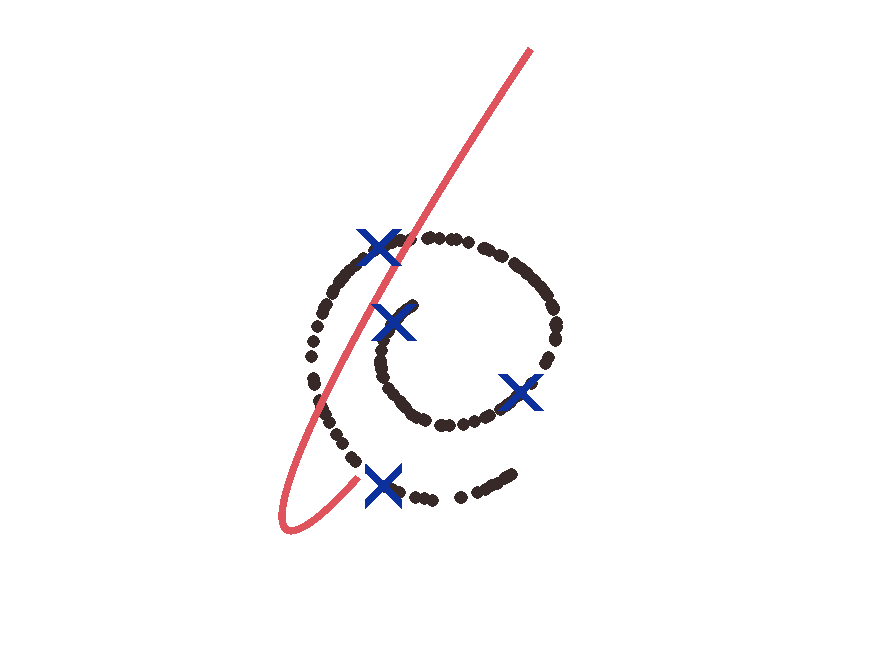}
	\caption{}
	\label{subfig:anchorInit_fail_1}
\end{subfigure}
\begin{subfigure}[b]{0.17\textwidth}
  \centering
	{\includegraphics[width=0.95\textwidth]{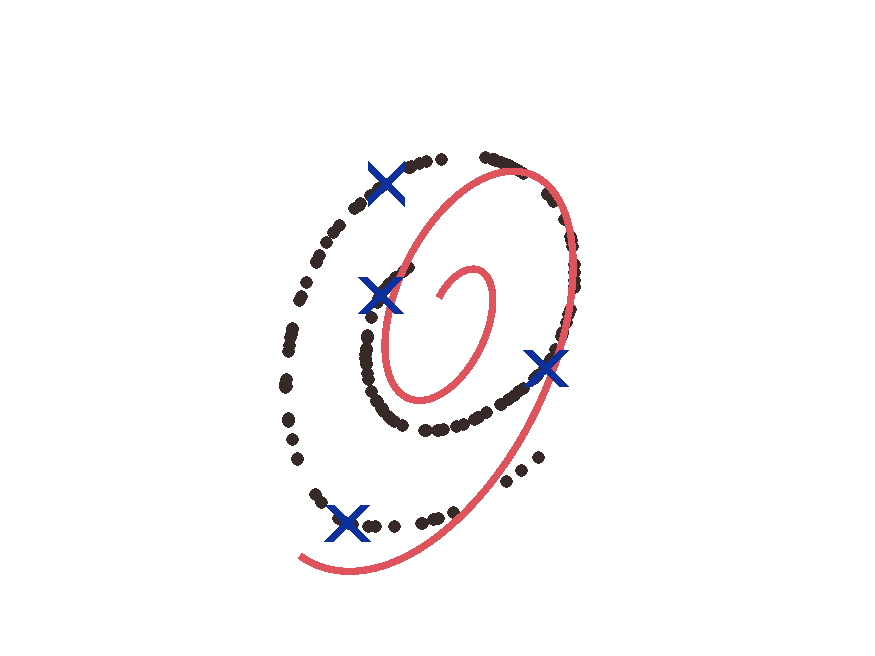}}
  \caption{}
	\label{subfig:anchorInit_fail_2}
\end{subfigure}
\begin{subfigure}[b]{0.17\textwidth}
  \centering
	{\includegraphics[width=0.95\textwidth]{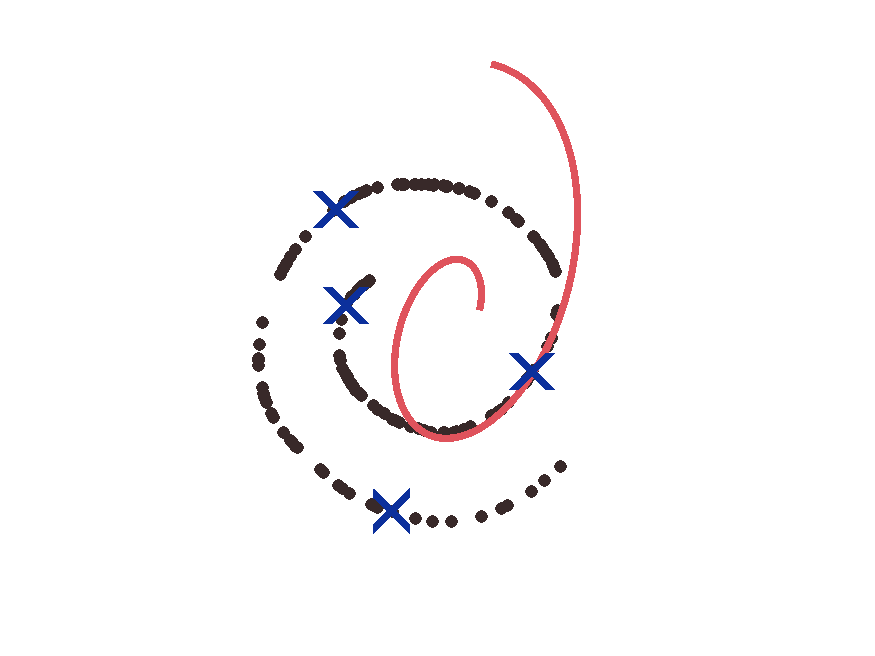}}
  \caption{}
	\label{subfig:anchorInit_fail_31}
\end{subfigure}

  \caption{\label{fig:anchorInit} Outputs of trials that are initialized with the same network weights and anchor point locations and different dictionary weights. The black dots are encoded data points, the blue x's are the encoded anchor point locations, and the red line is the generated orbit of the learned transport operator.(a-b) Trials that successfully learn a transport operator that traverses the swiss roll structure. (c-e) Trials that fail to learn a transport operator that traverses the swiss roll structure.}
	
\end{figure*}

Another possible factor important for successful learning of the latent manifold is the initialization of the dictionary elements. To analyze the impact of dictionary initialization, we train 10 separate instances of~\name{} which we initialize with the same network weights and anchor point locations and a different dictionary element weights. Fig.~\ref{fig:anchorInit} shows the final training outputs of five of these trials. The trials shown in Fig.~\ref{fig:anchorInit}(a-b) successfully learn a transport operator that traverses a swiss roll manifold and the trials shown in Fig.~\ref{fig:anchorInit}(c-e) fail to learn a transport operator that traverses the swiss roll manifold. This experiment shows that the final transport operator orbits and latent space encoding can vary significantly with different initializations of the dictionary element and this indicates that the dictionary initialization is a more important factor for successful training than the anchor point locations.

The transport operator objective is a non-convex optimization surface which can result in the optimization settling in local minima that do not represent the true data manifold. There are a few cues to observe when determining whether a training run is successfully learning the data manifold. First, as mentioned in Appendix~\ref{app:training}, during every transport operator update step, we check whether the update improves the portion of the objective that incorporates the transport operator. If this portion of the objective does not improve with a gradient step on the dictionary, we reject the step and decrease the learning rate. Often, if the initialized dictionary does not yield effective learning of the data manifold, then we see a large number of the gradient steps that do not improve the transport operator portion of the objective and those steps are rejected. Therefore, examining the number of rejected steps can be an indicator of the effectiveness of transport operator training. Another indicator for poor performance is the amount of time it takes to perform coefficient inference  when computing the prior. As the transport operator model becomes a better match for the latent manifold, the coefficient inference will require fewer steps which will reduce the amount of inference time. When comparing a \name{} model that learns to match the data manifold to one that is a poor match, the inference time is often noticeably lower for the successful model.

\section{CONCENTRIC CIRCLE EXPERIMENT}\label{app:concen_circle}

The concentric circle experiment uses the same network architecture as the swiss roll experiment which is specified in Table~\ref{tab:swissNet}. Table~\ref{tab:circleParams} shows the training parameters for the concentric circle experiment. In this experiment, we sample our ground truth 2D data manifold from two concentric circles and then map those sampled points to the 20-dimensional input space using a random linear mapping. We use 400 training points and randomly sample test points. 

It should be noted that, in this experiment, we do not alternate between steps where we update the network weights and anchor points while fixing the transport operator weights and steps where we update the transport operator weights while keeping the network weights and anchor points fixed. Instead the network weights, anchor points, and transport operator weights are all updated simultaneously. We use three anchor points per circular manifold and initialize them by evenly spacing them around each circle prior to mapping into the 20-dimensional input space. While the anchor points are allowed to update during training, the changes in the anchor points are negligible. For each input point, the prior is computed as a sum over the variational posterior conditioned on only the anchor points on the same circle as the input point.

\begin{table}[t]
\centering
\caption{Training Parameters for Concentric Circle Experiment}
\label{tab:circleParams}
\begin{tabular}{||l||} 
 \hline
 VAELLS Training - Concentric Circle  \\ 
 \hline
 batch size: 30   \\ 
 training steps: 4000   \\
 latent space dimension ($z_{dim}$): 2 \\
 $N_s: 1$ \\
 $lr_{\mathrm{net}}: 0.005$  \\
 $lr_{\mathrm{anchor}}: 0.0001$ \\
 starting $lr_{\Psi}: 4 \times 10^{-4}$  \\ 
 max $lr_{\Psi}: 0.1$ \\
 $\zeta_1:$ 0.01   \\
 $\zeta_2:$ 1   \\
 $\zeta_3:$ 1   \\
 $\zeta_4:$ 1   \\
 $\zeta_5:$ 0.01  \\
 $\zeta_q$: $1 \times 10^{-6}$\\
 $\zeta_p$:  $5 \times 10^{-6}$\\
 $\eta:$ 0.01   \\
 number of network and  updates steps: N/A  \\
 number of $\Psi$ update steps: N/A \\
 $\gamma_{\mathrm{post}}:$ 0.001\\
 warm-up steps: 0\\
 number of restarts for coefficient inference: 1 \\
 $M:$ 4\\
 number of anchors per class: 3\\
 latent space scale: 1 \\
 define prior with respect to closest anchor point: no \\
 \hline
\end{tabular}

\end{table}

\begin{figure}[t]
\centering

\includegraphics[width=0.99\textwidth]{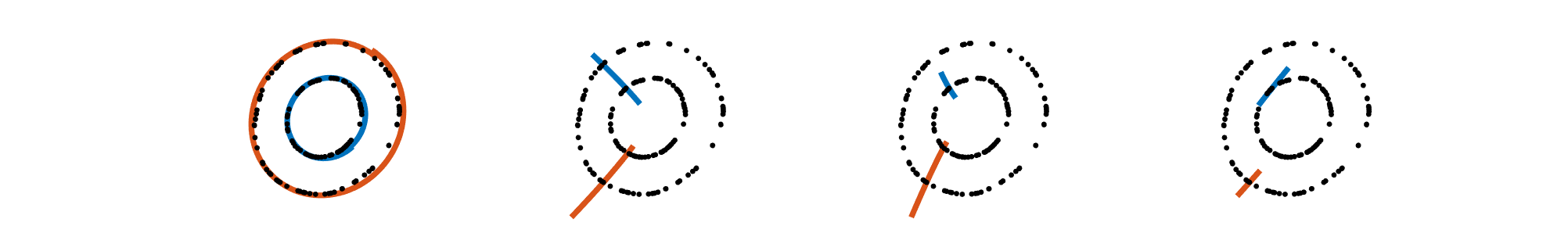}

	\caption{\label{fig:transOptOrbits} The orbits of each of the transport operators learned in the concentric circle test case plotted on top of encoded points.}

\end{figure}

\begin{figure}[t]
\centering

\includegraphics[width=0.3\textwidth]{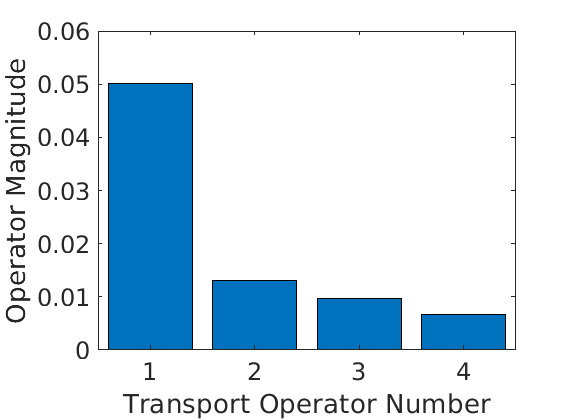}

	\caption{\label{fig:circleTOMag} Magnitude of the operators after training in the 2D concentric circle experiment.}

\end{figure}

Fig.~\ref{fig:transOptOrbits} shows the encoded latent points overlaid with the orbits of the learned transport operators. These orbits are generated by selecting one point on each circular manifold and applying a single operator as its trajectory evolves over time. Notice that one of the operators clearly represents the circular structure of the latent space while the other three have much smaller magnitudes and a limited effect on the latent space transformations. The Frobenius norm regularizer in the objective function often aids in model order selection by reducing the magnitudes of operators that are not used to represent transformations between points on the manifold. To see the magnitudes more clearly, Fig.~\ref{fig:circleTOMag} shows the magnitude of each of the transport operators after training the \name\ model in the concentric circle test case.

\begin{figure}

\centering
\begin{subfigure}[b]{0.26\textwidth}
 \centering
 {\includegraphics[width=0.95\textwidth]{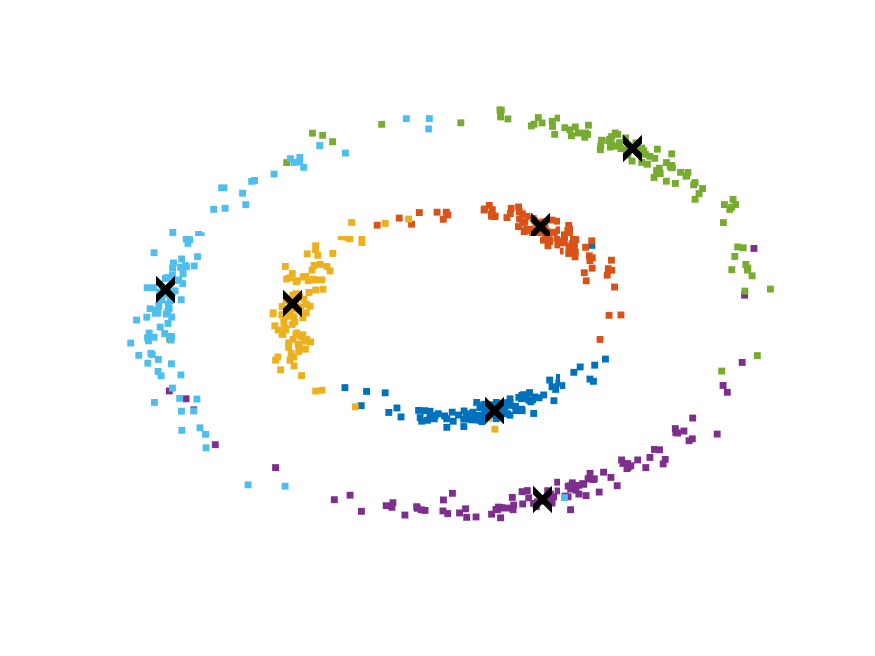}}
	\caption{}
	\label{subfig:circleSample} 
\end{subfigure}
\begin{subfigure}[b]{0.26\textwidth}
  \centering
	{\includegraphics[width=0.95\textwidth]{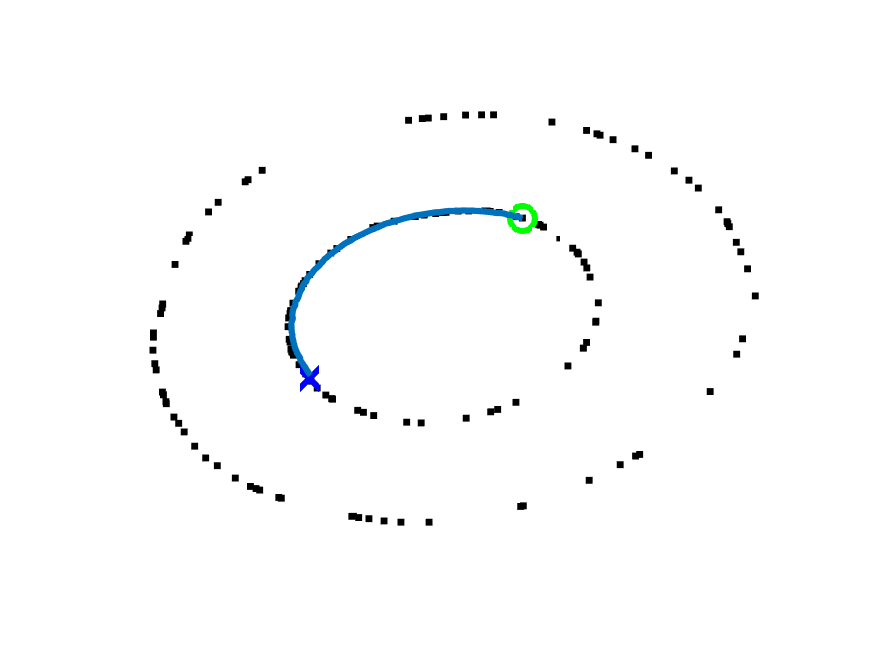}}
  \caption{}
	\label{subfig:inferPath_circle1}
\end{subfigure}
\begin{subfigure}[b]{0.26\textwidth}
  \centering
	{\includegraphics[width=0.95\textwidth]{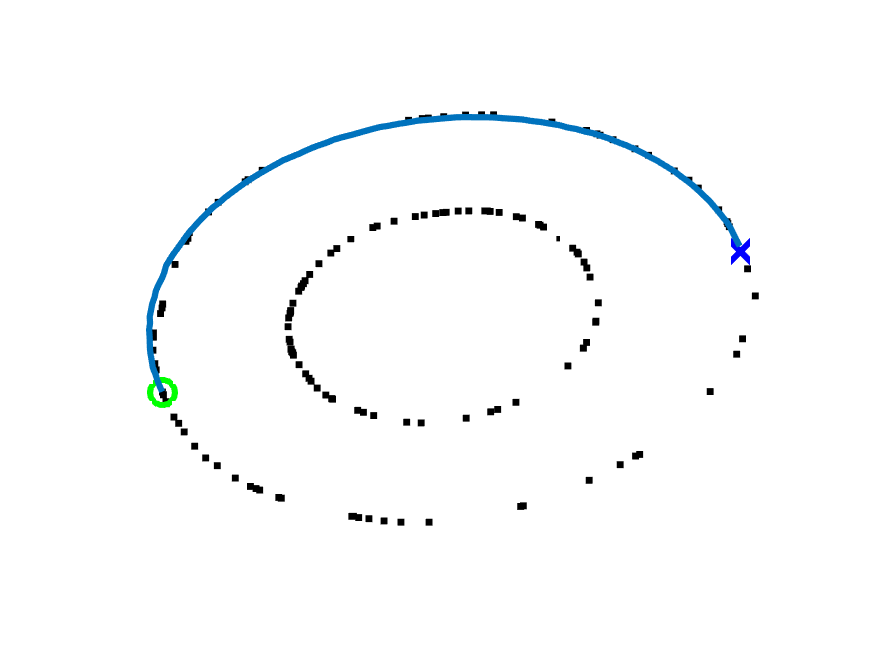}}
  \caption{}
	\label{subfig:inferPath_circle2}
\end{subfigure}
	
  \caption{\label{fig:inferredPath_circ} (a) Latent points sampled from the anchor points showing the concentric circle structure learned by the \name{}\ model. (b-c) Transport operator paths inferred between points on the same concentric circle manifold.}

\end{figure}

Fig.~\ref{subfig:circleSample} shows latent points sampled from the prior using the sampling described in \eqref{eq:sampling} with larger standard deviation and scale parameters to aid in visualization. Fig.~\ref{fig:inferredPath_circ}(b-c) show two example inferred paths between points encoded on the concentric circle manifold.

\section{ROTATED MNIST EXPERIMENT}\label{app:rot_mnist}

\begin{figure*}[ht]

\centering
\begin{subfigure}[b]{0.48\textwidth}
  \centering
	{\includegraphics[width=0.98\textwidth]{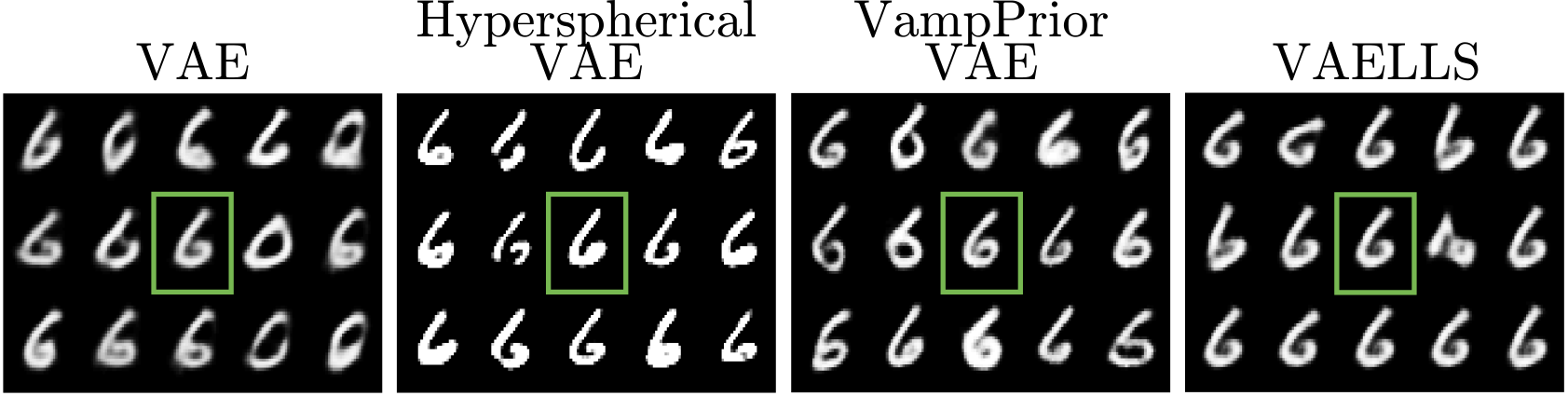}}
  \caption{}
	\label{subfig:rotDigit_Samp6}
\end{subfigure}
\begin{subfigure}[b]{0.48\textwidth}
  \centering
	{\includegraphics[width=0.98\textwidth]{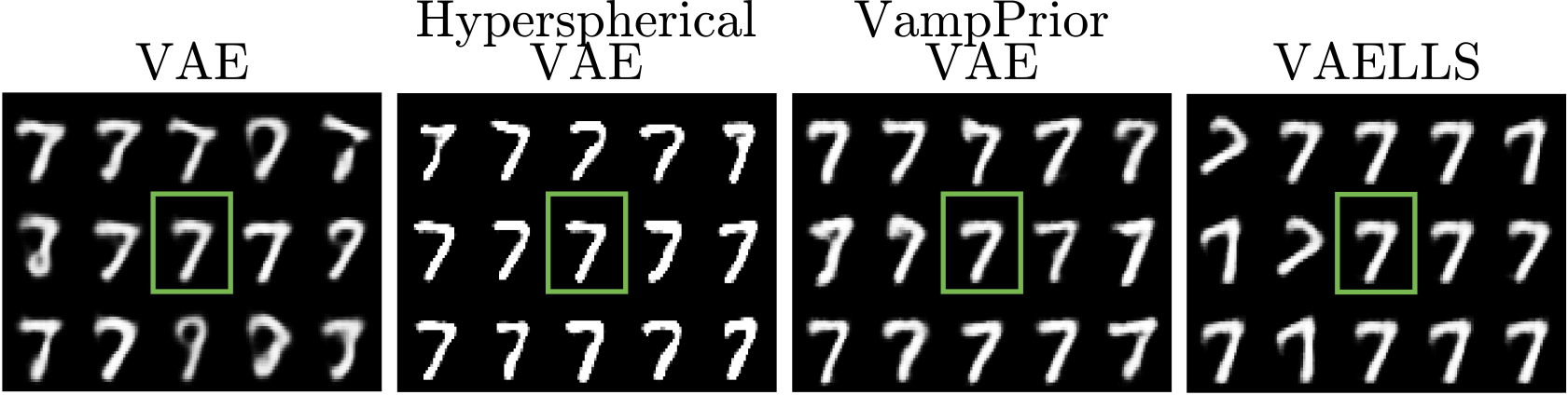}}
  \caption{}
	\label{subfig:rottDigit_Samp7}
\end{subfigure}

  \caption{\label{fig:rotDigit_samplingSupp} Examples of images decoded from latent vectors sampled from the posterior of models trained on rotated MNIST digits. In each example, the center digit (in the green box) is the decoded version of the input digit and the surrounding digits are images decoded from the sampled latent vectors. Sampling in the \name{}\ latent space results in rotations in the sampled outputs.}
	
\end{figure*}

We split the MNIST dataset into training, validation, and testing sets. The training set contains 50,000 images from the traditional MNIST training set. The validation set is made up of the remaining 10,000 image from the traditional MNIST training set. We use the traditional MNIST testing set for our testing set. The input images are normalized by 255 to keep the pixel values between 0 and 1. To generate a batch of rotated MNIST digits, we randomly select points from the MNIST training set and rotate those images to a random angle between 0 and 350 degrees. Separate anchor points are selected for each training example. Anchor points are generated by rotating the original MNIST sample by angles that are evenly spaced between 0 and 360 degrees. Because we have separate anchor points for each training sample, they are not updated during training. Tables~\ref{tab:rotMNISTNet} and~\ref{tab:rotMNISTParams} contain the VAELLS network architecture and parameters for the rotated MNIST experiment. As described in Appendix~\ref{app:training}, in this experiment, we define the prior only with respect to the anchor points that are estimated to be closest to each training point. Fig.~\ref{fig:rotDigit_samplingSupp} shows more examples of images decoded from latent vectors sampled from the posterior of the models trained on rotated MNIST digits. 

\begin{table}[t]
\centering
\caption{Network Architecture for MNIST Experiments}
\label{tab:rotMNISTNet}
\begin{tabular}{||l l||} 
 \hline
 Encoder Network & Decoder Network  \\ 
 \hline
 Input $\in \mathbb{R}^{28 \times 28}$ & Input $\in \mathbb{R}^2$  \\ 
 conv: chan: 64 , kern: 4, stride: 2, pad: 1  & Linear: 3136 Units  \\
 ReLU & ReLU \\
 conv: chan: 64, kern: 4, stride: 2, pad: 1  & convTranpose: chan: 64, kern: 4, stride: 1, pad: 1   \\ 
 ReLU & ReLU \\
 conv: chan: 64, kern: 4, stride: 1, pad: 0 & convTranpose: chann: 64, kern: 4, stride: 2, pad: 2   \\ 
 ReLU & ReLU \\
 Linear: 2 Units & convTranpose: chan: 1, kernel: 4, stride: 2, pad: 1  \\ 
   & Sigmoid \\ 
 \hline
\end{tabular}

\end{table}

\begin{table}[t]
\centering
\caption{Training Parameters for Rotated MNIST Experiment}
\label{tab:rotMNISTParams}
\begin{tabular}{||l||} 
 \hline
 VAELLS Training - Rotated MNIST \\ 
 \hline
 batch size: 32   \\ 
 training steps: 35000   \\
 latent space dimension ($z_{dim}$): 10 \\
 $N_s: 1$ \\
 $lr_{\mathrm{net}}: 10^{-4}$  \\
 $lr_{\mathrm{anchor}}: $ N/A \\
 starting $lr_{\Psi}: 1 \times 10^{-5}$  \\ 
 max $lr_{\Psi}: 0.008$ \\
 $\zeta_1:$ 1   \\
 $\zeta_2:$ 1   \\
 $\zeta_3:$ 1   \\
 $\zeta_4:$ 1   \\
 $\zeta_5:$ 0.01  \\
 $\zeta_q$: $1 \times 10^{-6}$\\
 $\zeta_p$:  $1 \times 10^{-6}$\\
 $\eta:$ 0.01   \\
 number of network and anchor update steps: 20  \\
 weight on prior terms during net update steps: 0.0001 \\
 number of $\Psi$ update steps: 60 \\
 weight on recon term during net update steps: 0.0001 \\
 $\gamma_{\mathrm{post}}:$ 0.001\\
 warm-up steps: 30000\\
 number of restarts for coefficient inference: 1 \\
 $M:$ 1\\
 number of anchors per class: 10\\
 latent space scale: 10 \\
 define prior with respect to closest anchor point: yes \\
 \hline
\end{tabular}

\end{table}

\section{NATURAL MNIST EXPERIMENT}\label{app:nat_mnist}

This experiment uses the same training/validation/testing separation as described in the rotated MNIST experiment in Section~\ref{app:rot_mnist}. The only pre-processing step for the digit images is normalizing by 255. Our qualitative results use a network that is trained with eight anchor points per digit class. These anchor points are initialized by randomly sampling eight examples of each class at the beginning of training. The anchor points are allowed to update during training but the changes in the anchor points during training are negligible. We use the same network architecture as in the rotated MNIST experiment (shown in Table~\ref{tab:rotMNISTNet}). Table~\ref{tab:natMNISTParams} shows the training parameters for this experiment.  As described in Appendix~\ref{app:training}, in this experiment, we define the prior only with respect to the anchor points that are estimated to be closest to each training point. 

\begin{table}[t]
\centering
\caption{Training Parameters for Natural MNIST Experiment}
\label{tab:natMNISTParams}
\begin{tabular}{||l||} 
 \hline
 VAELLS Training - MNIST \\ 
 \hline
 batch size: 32   \\ 
 training steps: 34600   \\
 latent space dimension ($z_{dim}$): 6 \\
 $N_s: 1$ \\
 $lr_{\mathrm{net}}: 10^{-4}1$  \\
 $lr_{\mathrm{anchor}}: 10^{-4}$ \\
 starting $lr_{\Psi}: 1 \times 10^{-5}$  \\ 
 max $lr_{\Psi}: 0.008$ \\
 $\zeta_1:$ 1   \\
 $\zeta_2:$ 1   \\
 $\zeta_3:$ 1   \\
 $\zeta_4:$ 1   \\
 $\zeta_5:$ 0.01  \\
 $\zeta_q$: $1 \times 10^{-6}$\\
 $\zeta_p$:  $1 \times 10^{-6}$\\
 $\eta:$ 0.01   \\
 number of network and anchor update steps: 20  \\
 weight on prior terms during net update steps: 0.0001 \\
 number of $\Psi$ update steps: 60 \\
 weight on recon term during net update steps: 0.0001 \\
 $\gamma_{\mathrm{post}}:$ 0.001\\
 warm-up steps: 30000\\
 number of restarts for coefficient inference: 1 \\
 $M:$ 4\\
 number of anchors per class: 8\\
 latent space scale: 10\\
 define prior with respect to closest anchor point: yes \\
 \hline
\end{tabular}

\end{table}

\begin{figure*}[ht]

\centering
\begin{subfigure}[b]{0.48\textwidth}
  \centering
	{\includegraphics[width=0.98\textwidth]{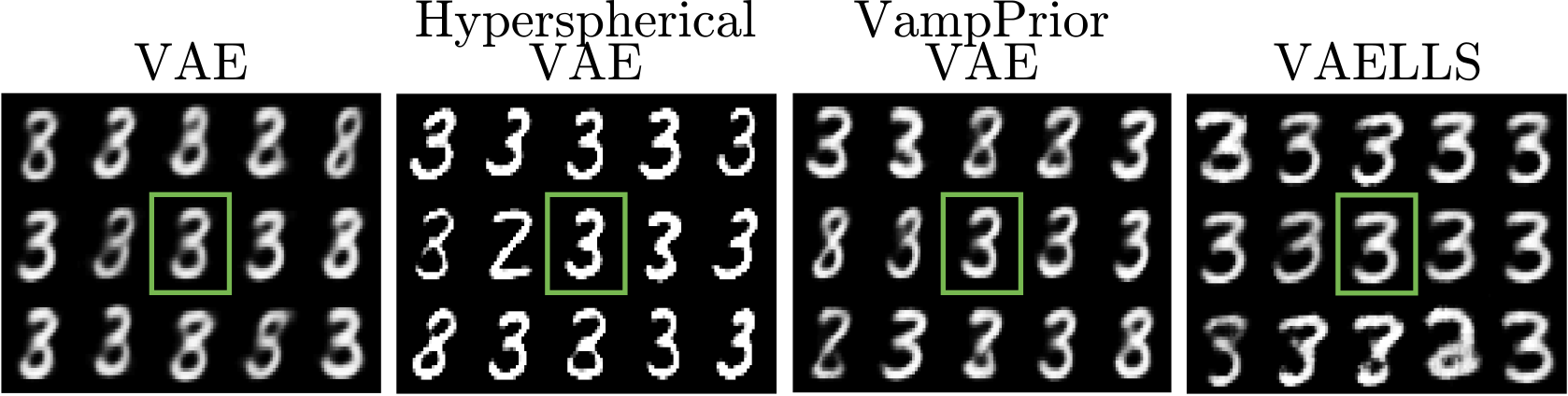}}
  \caption{}
	\label{subfig:natDigit_Samp3}
\end{subfigure}
\begin{subfigure}[b]{0.48\textwidth}
  \centering
	{\includegraphics[width=0.98\textwidth]{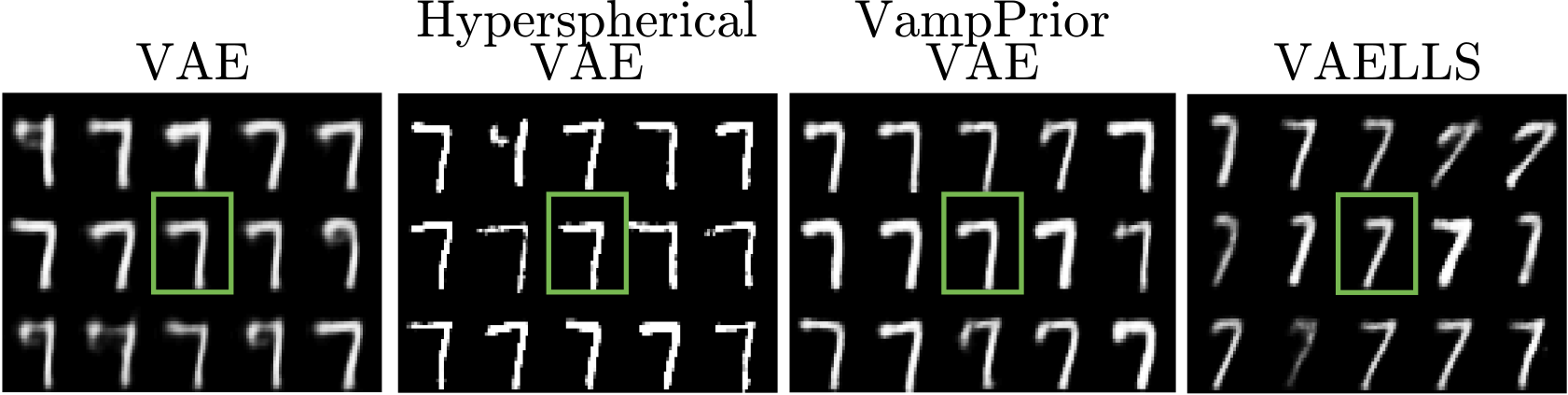}}
  \caption{}
	\label{subfig:natDigit_Samp9}
\end{subfigure}

  \caption{\label{fig:natDigit_samplingSupp} Examples of images decoded from latent vectors sampled from the posterior of models trained on natural MNIST digits. In each example, the center digit (in the green box) is the decoded version of the input digit and the surrounding digits are images decoded from the sampled latent vectors.}
	
\end{figure*}

\begin{figure*}[ht]

\centering
\begin{subfigure}[b]{0.4\textwidth}
  \centering
	{\includegraphics[width=0.98\textwidth]{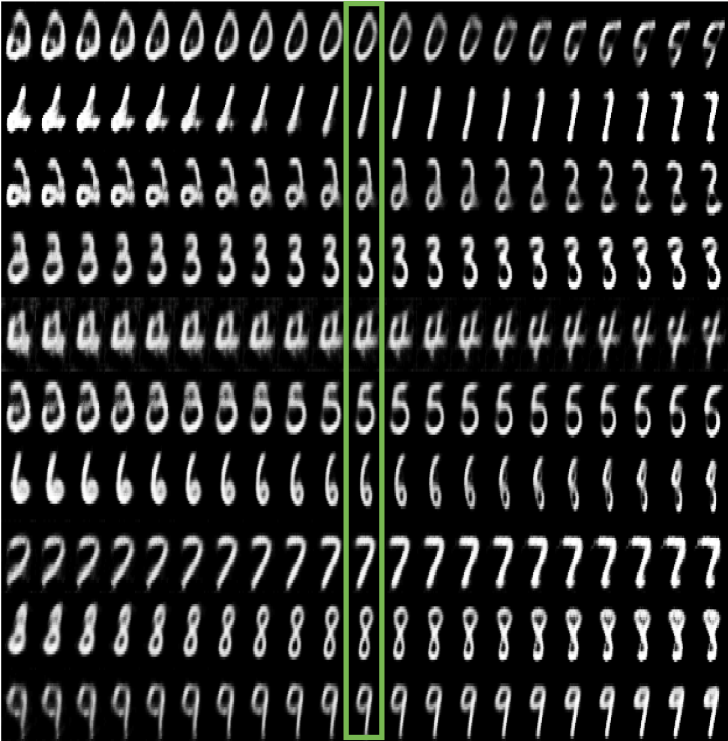}}
  \caption{}
	\label{subfig:natDigit_TO1}
\end{subfigure}
\begin{subfigure}[b]{0.4\textwidth}
  \centering
	{\includegraphics[width=0.98\textwidth]{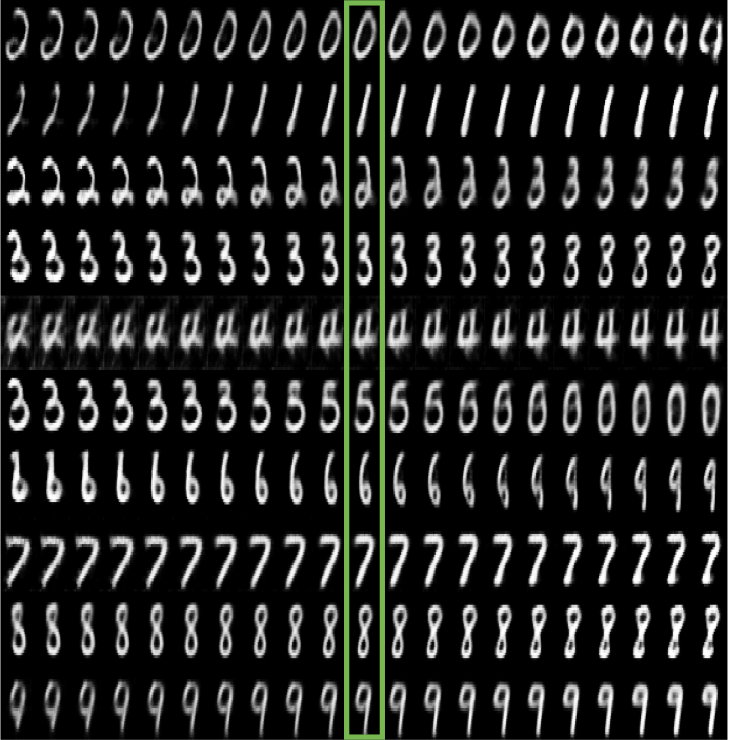}}
  \caption{}
	\label{subfig:natDigit_TO2}
\end{subfigure}
\begin{subfigure}[b]{0.4\textwidth}
  \centering
	\includegraphics[width=0.98\textwidth]{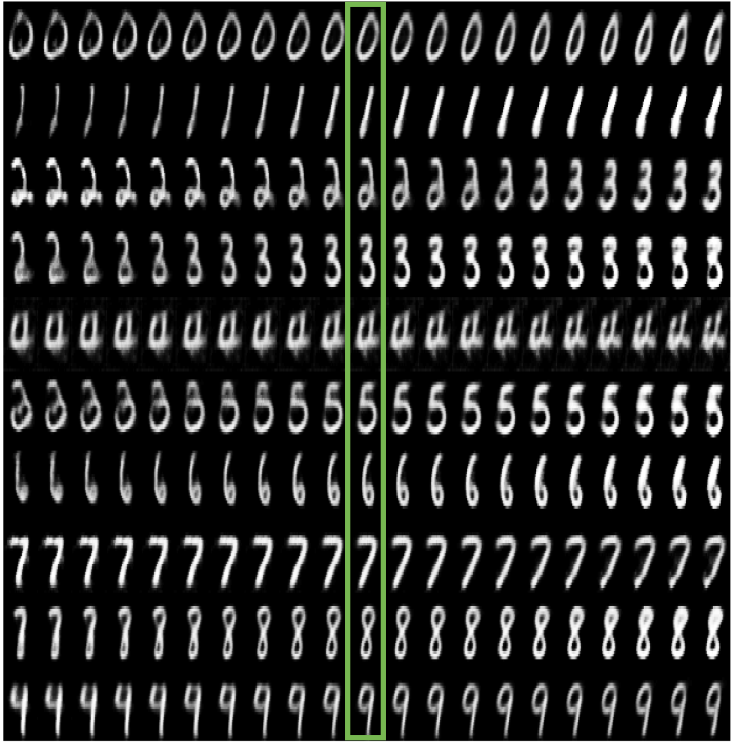}
	\caption{}
	\label{subfig:natDigit_TO3}
\end{subfigure}
\begin{subfigure}[b]{0.4\textwidth}
  \centering
	{\includegraphics[width=0.98\textwidth]{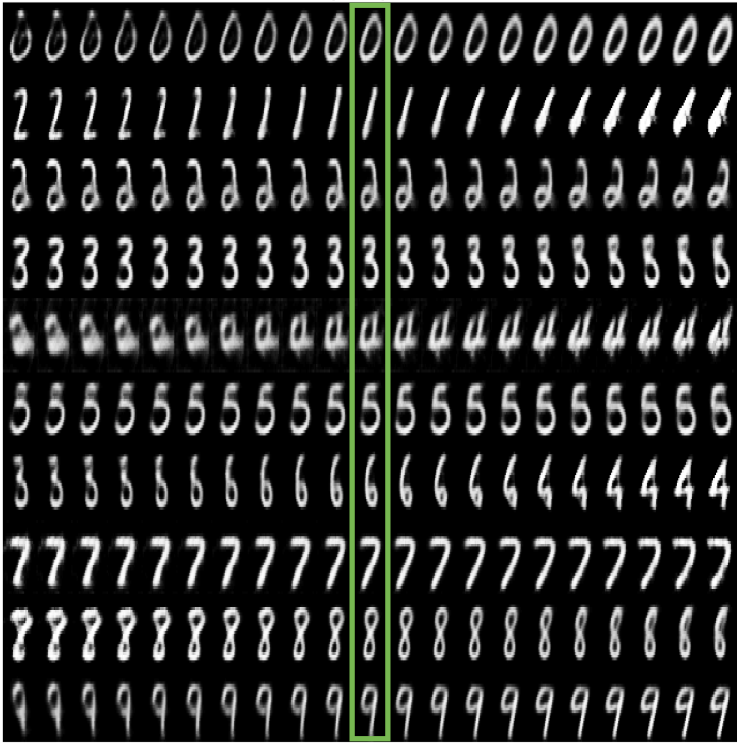}}
  \caption{}
	\label{subfig:natDigit_TO4}
\end{subfigure}

  \caption{\label{fig:natDigit_TOAll1} Extrapolated paths using the first four  transport operators learned on natural MNIST digit variations.}
	
\end{figure*}

\begin{figure*}[ht]

\centering
\begin{subfigure}[b]{0.4\textwidth}
  \centering
	{\includegraphics[width=0.98\textwidth]{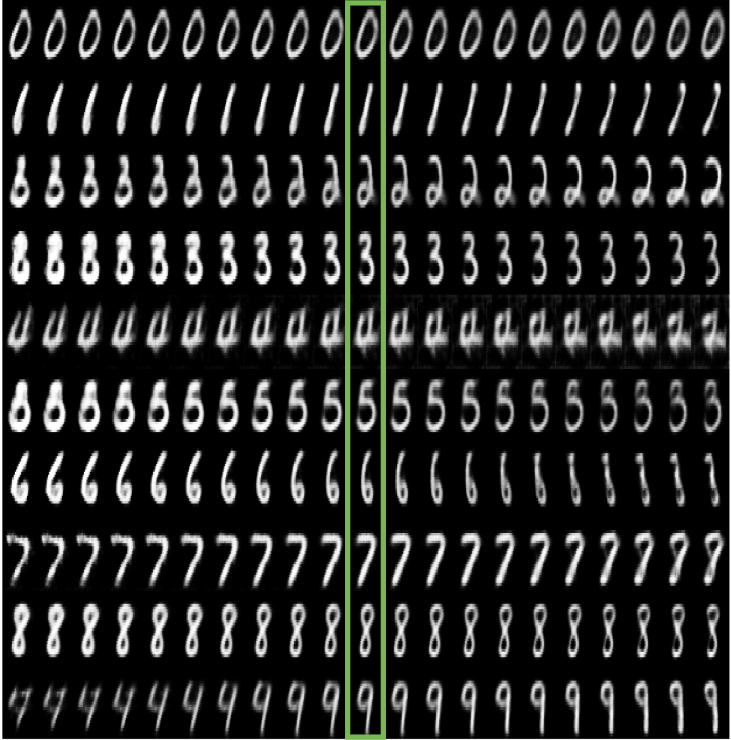}}
  \caption{}
	\label{subfig:natDigit_TO5}
\end{subfigure}
\begin{subfigure}[b]{0.4\textwidth}
  \centering
	{\includegraphics[width=0.98\textwidth]{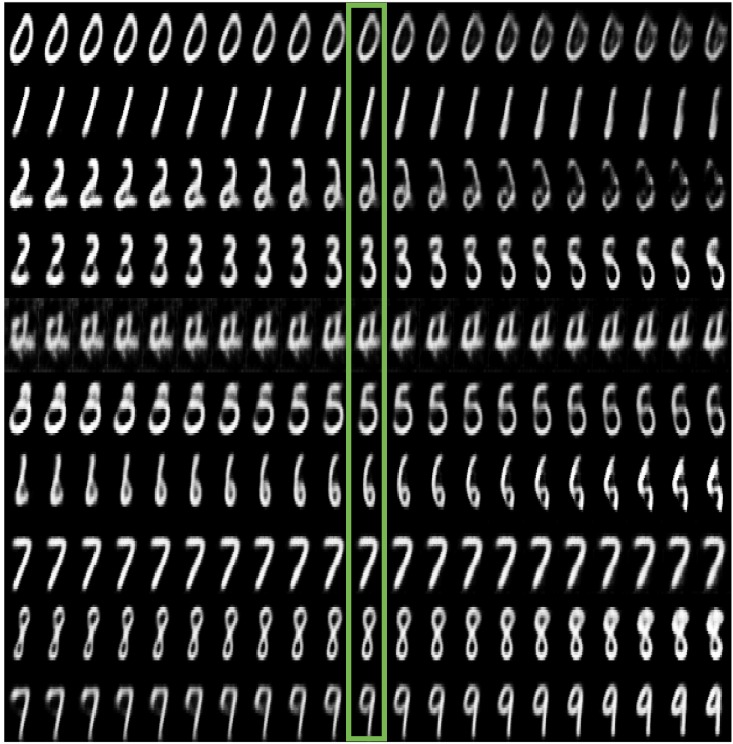}}
  \caption{}
	\label{subfig:natDigit_TO6}
\end{subfigure}
\begin{subfigure}[b]{0.4\textwidth}
  \centering
	\includegraphics[width=0.98\textwidth]{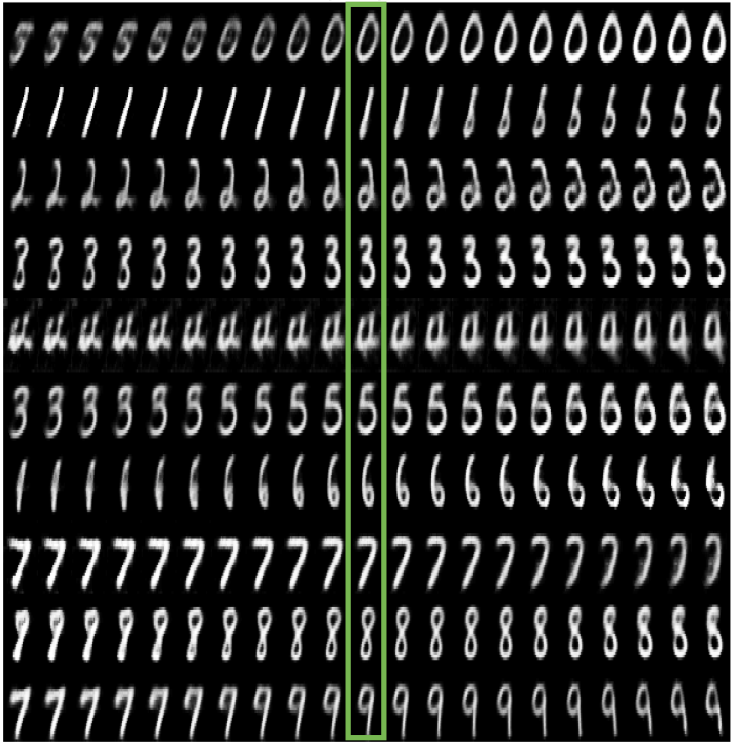}
	\caption{}
	\label{subfig:natDigit_TO7}
\end{subfigure}
\begin{subfigure}[b]{0.4\textwidth}
  \centering
	{\includegraphics[width=0.98\textwidth]{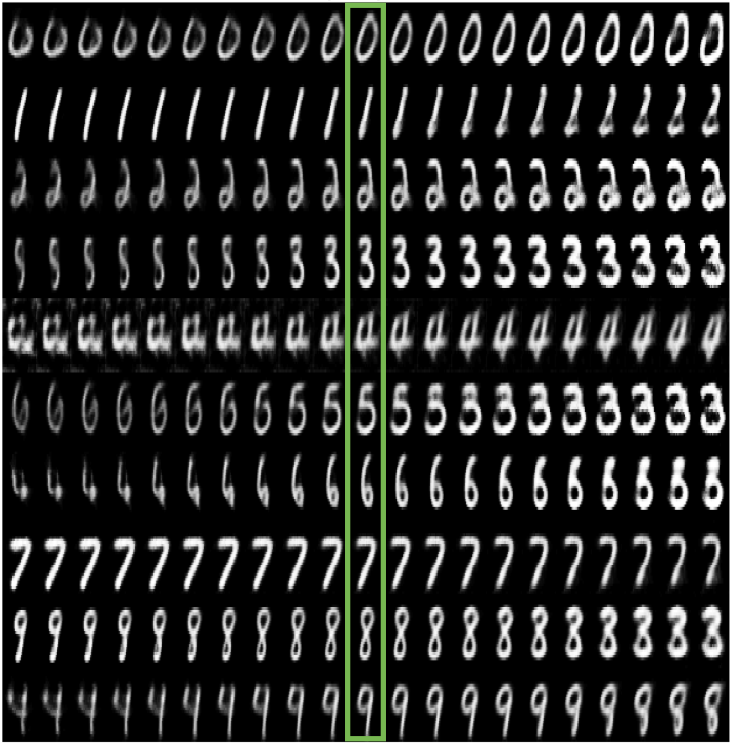}}
  \caption{}
	\label{subfig:natDigit_TO8}
\end{subfigure}

  \caption{\label{fig:natDigit_TOAll2} Extrapolated paths using the second four transport operators learned on natural MNIST digit variations.}
	
\end{figure*}

Fig.~\ref{fig:natDigit_samplingSupp} shows more examples of images decoded from latent vectors sampled from the posterior of the models trained on MNIST digits. Fig.~\ref{fig:natDigit_TOAll1} and Fig.~\ref{fig:natDigit_TOAll2} show the effect that each of the eight learned transport operators has on digits. To generate each figure, input images randomly selected from each class are encoded into the latent space and a single learned operator is applied to each of those latent vectors. The decoded version of the input image is shown in the middle column (in a green box). The images to the left of the middle column show the result of applying the operator with a negative coefficient and the images to the right of the middle column show the result of applying the operator with a positive coefficient.

\end{document}